\documentclass[11pt]{article}

\usepackage[preprint]{acl}

\usepackage{times}
\usepackage{latexsym}

\usepackage[T1]{fontenc}
\usepackage[utf8]{inputenc}

\usepackage{microtype}

\usepackage{inconsolata}

\usepackage{graphicx}

\usepackage{color}
\usepackage{soul}
\usepackage{amsmath}
\usepackage{amsfonts}
\usepackage{amssymb}
\usepackage{graphicx}
\usepackage{enumitem}
\usepackage{multirow}
\usepackage{array}
\usepackage{float}
\usepackage{adjustbox}
\usepackage{pifont}
\usepackage{multicol}
\usepackage{booktabs}
\usepackage{makecell}
\usepackage{xspace}
\usepackage{tcolorbox}
\usepackage[normalem]{ulem}
\usepackage{cleveref}
\usepackage{wrapfig}
\usepackage{xcolor}
\usepackage{colortbl}
\usepackage{arydshln}
\usepackage{tabularx}
\usepackage{afterpage}
\usepackage{caption}
\usepackage{subcaption}
\usepackage{etoolbox}
\usepackage{algorithm}
\usepackage{algpseudocode}

\usepackage{tcolorbox}
\newtcolorbox{verbatimbox}{
  colback=black!4, % Background color
  colframe=black!4, % Border color
  arc=1pt, % Radius of the rounded corners
  boxrule=0.5pt, % Border width
  fontupper=\ttfamily, % Use monospaced font for verbatim text
  left=2pt, right=2pt, % inner margin
}
\AtBeginEnvironment{verbatimbox}{\fontsize{9}{9}\selectfont}

\renewcommand{\cite}{\citep}

\setlist[itemize]{itemsep=0.02cm,topsep=0.2cm,left=0cm}
\setlist[enumerate]{itemsep=0.02cm,topsep=0.2cm,left=0cm}

\definecolor{syntaxper}{HTML}{309f04}
\definecolor{syntaxloc}{HTML}{C9AB40}
\definecolor{syntaxkeyword}{HTML}{777777}
\definecolor{syntaxtag}{HTML}{000000}
\definecolor{syntaxnum}{HTML}{700156}

\newcommand{\hlper}[1]{\textcolor{syntaxper}{\textbf{#1}}}
\newcommand{\hlloc}[1]{\textcolor{syntaxloc}{\textbf{#1}}}

\newcommand{\hlkw}[1]{\textcolor{syntaxkeyword}{\textbf{#1}}}
\newcommand{\hltag}[1]{\textcolor{syntaxtag}{#1}}
\newcommand{\hlnum}[1]{\textcolor{syntaxnum}{#1}}

\newcommand{\cattag}{\textit{tagging}}
\newcommand{\catmatch}{\textit{matching}}
\newcommand{\catindex}{\textit{indexing}}
\newcommand{\Cattag}{\textit{Tagging}}
\newcommand{\Catmatch}{\textit{Matching}}
\newcommand{\Catindex}{\textit{Indexing}}
\newcommand{\ourmethod}{LogitMatch}

\newcommand{\hlperx}[1]{{\leavevmode\color{syntaxper} {\ul{\textbf{#1}}}}}
\newcommand{\hllocx}[1]{{\leavevmode\color{syntaxloc} {\ul{\textbf{#1}}}}}

\title{Strategies for Span Labeling with Large Language Models}
\author{
  \begin{tabular}{ccc}
    {\bf Danil Semin} & {\bf Ondřej Dušek} & {\bf Zdeněk Kasner} \\
    \texttt{semin@ufal.mff.cuni.cz} & \texttt{odusek@ufal.mff.cuni.cz} & \texttt{kasner@ufal.mff.cuni.cz}
  \end{tabular} \\[0.5em]
  Institute of Formal and Applied Linguistics \\
  Faculty of Mathematics and Physics, Charles University
}
         
\begin{document}
\maketitle
\begin{abstract}
Large language models (LLMs) are increasingly used for text analysis tasks, such as named entity recognition or error detection. Unlike encoder-based models, however, generative architectures lack an explicit mechanism to refer to specific parts of their input. This leads to a variety of ad-hoc prompting strategies for input text span labeling, often with inconsistent results. In this paper, we categorize these strategies into three families: \textit{tagging} the whole input text, \textit{indexing} numerical positions of spans, and \textit{matching} generated span content. To address the limitations of content matching, we introduce \textsc{LogitMatch}, a new constrained decoding method that forces the model's output to align with valid input spans. On four diverse tasks, we find that tagging-based methods are the most robust, but \textit{matching} methods are competitive and more token-efficient. Our \textsc{LogitMatch} approach improves upon \emph{matching} methods by eliminating span matching issues while retaining their performance.\footnote{Our code is available at \url{https://github.com/semindan/span_labeling}.}
\end{abstract}

% \footnote{Our code is available at \url{https://anonymous.4open.science/r/llms-7677}.}

\section{Introduction}
Generative large language models (LLMs) are mostly based on the Transformer decoder architecture \cite{vaswani2017attention,radford2019language}, which makes them well-equipped for open-ended generative tasks. However, LLMs are increasingly used for text analysis tasks as well: from finding factual inconsistencies or grammatical errors to extracting structured information about events and entities. In these tasks, an issue arises: \textbf{how can LLMs explicitly refer to parts of their input text?}

\begin{figure}[t]
    \centering
    \includegraphics[width=\columnwidth]{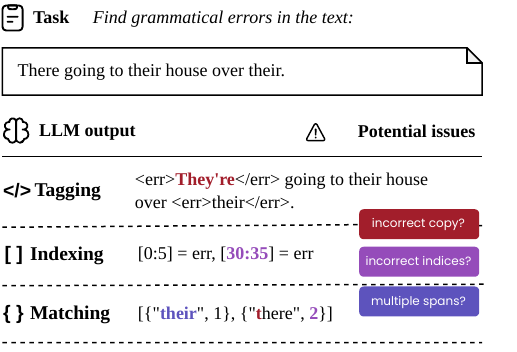}

  \caption{Main span labeling strategies: tagging, matching, and indexing. The issues include inability to copy the input text in verbatim, wrong numerical indices, and ambiguities stemming from multiple identical spans. }
  %Given an input text, the goal is to identify spans in the text by generating textual output.
  \label{fig:teaser}
\end{figure}

While constraining the model output structure is a commonly supported feature in LLM inference frameworks \cite{liu2024we,willard2023efficient,guidance}, there is no straightforward way to constrain the model output to contain only parts of the given text. Unlike encoder-based models \cite{devlin2019bertpretrainingdeepbidirectional,warner2024smarterbetterfasterlonger} that can attach a label to each input token, generative LLMs, by contrast, have no explicit mechanism to refer to their input. This issue becomes apparent in span labeling tasks, where we need to identify and categorize parts of the given text (see \Cref{fig:teaser} for an example). In these cases, we find ourselves needing to \emph{refer to spans in the given input text by generating another text}, which offers no guarantees on alignment between the two texts.

%The naive strategy -- prompting the model to produce the output with the given constraints -- may often fail, leading to parsing errors and degraded performance on the task \cite{long2024llms,macedo2024exploring,molfese2025right}. 

% https://drive.google.com/file/d/19cDfjukSmCNrS70eTnwVfHvuYXdRyuCo/view?usp=sharing

Why use LLMs for span labeling? First, LLMs can achieve similar performance on span labeling tasks as finetuned encoder models without costly task-specific training \cite{wang2023gptnernamedentityrecognition,kocmi2024error}. Thanks to their in-context learning abilities, LLMs can also deal with low-resource tasks for which encoder models are not readily available, such as evaluating factual accuracy of sports reports \cite{thomson2023evaluating}, identifying rhetorical structures \cite{mulcaire2025span}, or detecting fallacies and propaganda techniques \cite{hasanain2024large,ramponi2025finegrained}.

In this paper, we explore how to robustly apply LLMs to span labeling tasks. We first show that there is currently no consensus on how to formulate span labeling tasks for LLMs. This leads to a wide range of ad hoc strategies, often found to deliver unsatisfactory performance on downstream tasks. In \Cref{sec:methods}, we cluster these strategies into three groups: tagging a copy of the whole input text (\cattag{}), referring to the numerical indices of the input tokens (\catindex{}), and matching the generated span content (\catmatch{}). 

% On top of the base methods for each strategy, we also present tweaks that were introduced to address their drawbacks.

In \Cref{sec:our_method}, we introduce our novel constrained decoding extension for the \catmatch{} strategies. Our method -- \textsc{\ourmethod{}} -- provides guarantees on the output format by modifying raw model logits to only allow generating valid spans from the input text. By affecting only the decoding process, our method is applicable to any locally-deployed LLM without costly model finetuning or architecture modifications.

In \Cref{sec:tasks}, we evaluate a representative set of span labeling methods across four diverse tasks: named entity recognition, grammatical error correction, machine translation error detection, and a synthetic pattern lookup task. In \Cref{sec:experiments}, we discuss our results, showing that while \cattag{} strategies offer a robust baseline, \textsc{\ourmethod{}} successfully addresses the limitations of standard \catmatch{} strategies and offers a highly competitive approach. We also analyze the trade-offs of each method and provide recommendations regarding their applications to specific tasks.

\section{Task Definition}
\label{sec:task_definition}

In this paper, our goal is to (1) \textbf{identify spans}, i.e., contiguous sequences in an input text satisfying given criteria and (2) \textbf{classify each span} into one of predefined categories. Both of these subtasks need to be accomplished by generating textual output, in order to apply LLMs without architecture modifications.

% \footnote{The concept is equivalent to token classification (as known, e.g., from part-of-speech tagging) if we assume that all tokens outside of spans belong to a default \emph{unlabeled} category.}

Formally, we define the problem as follows: Given an input text $X = [x_1, x_2, \ldots, x_n]$ consisting of $n$ characters, the objective is to identify and label a set of spans $S = \{(b_i, e_i, c_i)\}_{i=1}^m$ in $X$ satisfying given criteria. Each span is characterized by three components: a begin index $b_i$, an end index $e_i$ (exclusive), and a category $c_i$ from the set of span categories $C$ defined for the task. The begin and end indices satisfy $1 \leq b_i \leq e_i \leq n + 1$. In this work, we consider the spans to be non-overlapping. We are using a model $\mathcal{M}$ that generates an output string $Y = [y_1, y_2, \ldots, y_{|Y|}]$, out of which $S$ needs to be extracted.

\section{Existing Strategies for Span Labeling with Generative Models}
\label{sec:methods}

\begin{table*}[t]
\centering
\small
\begin{tabular}{l l >{\raggedright\arraybackslash\microtypesetup{protrusion=false}}p{8cm} p{4cm}}
\multicolumn{4}{l}{\rule{0pt}{2ex}\textbf{Target output}: \hlperx{Turing \textsuperscript{PER}} was born in \hllocx{London \textsuperscript{LOC}}.}   \\
\addlinespace[3pt]
\midrule
\textbf{Str.} & \textbf{Method} & \textbf{Output} & \textbf{Used in} \\
\midrule
\multirow{7}{*}[-1.2em]{\rotatebox{90}{\cattag{}}} 
& \multirow{3}{*}{Custom tags}
& \texttt{\hltag{@@}Turing\hltag{\#\#} was born in \hltag{@@}London\hltag{\#\#}.} 
& \citet{wang2023gptner}   \\
\addlinespace[2pt]
\cdashline{3-4}[.4pt/1pt] 
\addlinespace[2pt]
& & \texttt{\hltag{@@}Turing\hltag{\#\#}\hlper{PER} was born in \hltag{@@}London\hltag{\#\#}\hlloc{LOC}.}
& \citet{yan2024ltner} \\
\cmidrule(lr){2-4}
& XML tags
& \texttt{\hltag{<entity type=}"\hlper{PER}"\hltag{>}Turing\hltag{</entity>} was born in \newline\hspace*{0.5em}\hltag{<entity type=}"\hlloc{LOC}"\hltag{>}London\hltag{</entity>}\hltag{.}}
& \citet{treviso2024xtower}, \newline \citet{mulcaire2025span} \\
\cmidrule(lr){2-4}
& \multirow{3}{*}{BIO tags}
& \texttt{(Turing, \hlper{B-PER}) (was, \hlkw{O}) (born, \hlkw{O}) \newline \hspace*{0.5em}(in, \hlkw{O}) (London, \hlloc{B-LOC})} 
& \citet{obeidat2025llms}, \newline \citet{ramponi2025finegrained} \\
\addlinespace[2pt]
\cdashline{3-4}[.4pt/1pt] 
\addlinespace[2pt]
& & \texttt{\hlper{B-PER} \hlkw{O} \hlkw{O} \hlkw{O} \hlloc{B-LOC}}
&  \citet{wang2023gptner} \\
\midrule
\multirow{4}{*}[-0.9em]{\rotatebox{90}{\catindex{}}} 
& Char-level
& \texttt{[\hlnum{1}:\hlnum{7}] = \hlper{PER}}\newline \texttt{[\hlnum{20}:\hlnum{26}] = \hlloc{LOC}} 
& \citet{hasanain2024large} \\
\cmidrule(lr){2-4}
& Word+num
& \textit{Input: [1]Turing [2]was [3]born [4]in [5]London} \newline 
\texttt{[\hlnum{1}:\hlnum{1}] = \hlper{PER}}\newline \texttt{[\hlnum{5}:\hlnum{5}] = \hlloc{LOC}} & \citet{ramponi2025finegrained} \\
\midrule
\multirow{5}{*}[-1em]{\rotatebox{90}{\catmatch{}}} 
& Bullet points
& \texttt{- \hlper{PER} - Turing} \newline \texttt{- \hlloc{LOC} - London}
& \citet{kocmi2023gembamqm} \\
\cmidrule(lr){2-4}
& \multirow{3}{*}{JSON output}
& \texttt{[\{"type": "\hlper{PER}", "text": "Turing"\},} \newline \hspace*{0.5em}\texttt{\{"type": "\hlloc{LOC}", "text": "London"\}]} 
& \citet{kasner2024traditional}, \newline \citet{kasner2025large} \\
\addlinespace[2pt]
\cdashline{3-4}[.4pt/1pt] 
\addlinespace[2pt]
& & \texttt{[\{"type": "\hlper{PER}", "text": "Turing", "index": \hlnum{0}\},} \newline \hspace*{0.5em}\texttt{\{"type": "\hlloc{LOC}", "text": "London", "index": \hlnum{0}\}]} 
& \citet{klesnilova2025multilingual}, \newline \citet{hasanain2024large} \\
\bottomrule
\end{tabular}
\caption{Overview of span labeling strategies with generative models used in related work. As an example, we present the methods applied on the NER task for the sentence ``\emph{Turing was born in London.}'' with two span categories: \texttt{\hlper{PER}} (person) and \texttt{\hlloc{LOC}} (location).}
\label{tab:methods_overview}
\end{table*}

We identified three groups of strategies in the literature that are used to formulate span labeling for generative models (see \Cref{tab:methods_overview} for an overview). We categorize these strategies depending on how the spans are marked: \cattag{} (\Cref{sec:copy}), \catindex{} (\Cref{sec:numerical}), and \catmatch{} (\Cref{sec:json}).

\subsection{\Cattag{}: Marking Spans in the Input Text}
\label{sec:copy}

\Cattag{} strategies explicitly generate an output for each input token, requiring $O(n)$ output tokens per example. In the more common variant, the model generates a full copy of the input text and surrounds the spans of interest with special tags. Alternatively, the model can generate a sequence of tags, one tag per input token, and include \textit{null} tags for unmarked tokens. 

\paragraph{Custom Tags} One of the first applications of LLMs for span labeling is the work of \citet{wang2023gptner}, who apply an LLM to identifying named entities in a text. They prompt the model to surround the entities with special markers \texttt{@@} and \texttt{\#\#} to denote tag boundaries. \citet{yan2024ltner} extend their approach by instructing the model to append the entity category after the closing tag (such as \texttt{\#\#LOC}), adding the possibility to classify the spans.

\paragraph{XML-like Tags} To detect translation errors, \citet{treviso2024xtower} instruct an LLM to surround errors with XML-like tags such as \texttt{<error1 severity="major">{text}</error1>}. A similar approach is adopted by \citet{mulcaire2025span} to detect shell language (filler words and phrases) in English language exams.

\paragraph{BIO Tags} The BIO scheme \cite{ramshaw-marcus-1995-text} classifies each token as a beginning of a span (\texttt{B}), inside a span (\texttt{I}), or outside a span (\texttt{O}). \citet{obeidat2025llms} use the BIO scheme for biomedical NER by making the LLM generate a pair \texttt{word:tag} for all  words in the input text. The BIO tagging scheme was also initially tested by \citet{ramponi2025finegrained} to detect errors in social media texts, but it was discarded due to unsatisfactory results. An alternative approach proposed by \citet{wang2023gptner} also relies on BIO tags but only generates the sequence of tags without generating the words themselves. However, this underperforms their aforementioned custom tag approach.

\subsection{\Catindex{}: Referring to Character Indices}
\label{sec:numerical}
\Catindex{} strategies output spans to be labeled, which requires $O(m)$ tokens per example (i.e., grows linearly with the number of annnotated spans). 
%Unlike \catmatch{} approaches, they bypass string matching of the span content by 
The model is explicitly asked for the numerical begin and end indices ($b_i$, $e_i$) of each span.

\paragraph{Character-level} For propaganda span detection, \citet{hasanain2024large} ask for the start and end character indices of the span. They report that the method led to inaccuracies since the model did not have explicit access to character indices. The authors eventually resorted to string matching of the span content (cf. \Cref{sec:json}).

\paragraph{Word-level with numbering} For the task of detecting argumentation fallacies, \citet{ramponi2025finegrained} provide an LLM with explicit access to word indices by appending an index to each word in the input text.
% , e.g., ``\texttt{[1]Turing [2]was...}'' as illustrated in \Cref{tab:methods_overview}. 
The output they require is in the format \texttt{[start\_word:end\_word = category]}. This approach was found to perform better than using BIO tags (cf. \Cref{sec:copy}).

\subsection{\Catmatch{}: String Matching of the Span Content}
\label{sec:json}
\Catmatch{} strategies generate the textual content of each span. As such, they generate $O(m \cdot \bar{l})$ tokens per example, where $\bar{l}$ is the average length of span content.\footnote{Since the span content can cover the entire input text in the extreme, the upper bound is $O(n)$. However, the spans tend to be much shorter in practice.} Unlike in \catindex{} strategies, the model generates the textual content of the spans. The begin and end indices ($b_i$, $e_i$) are determined post-hoc by string matching. 

\paragraph{Unstructured List} A basic version of this approach is adopted by \citet{kocmi2023gembamqm}, who detect translation errors by asking the model to generate a bulleted list with items \texttt{\{error\_category\} - \{span\_text\}}.

\paragraph{JSON Output} \citet{kasner2024traditional} use a JSON-like format for evaluating output accuracy in data-to-text generation. The approach is also applied by \citet{kasner2025large} to identify errors in machine translation and detect propaganda techniques. Similarly, \citet{hasanain2024large} and \citet{klesnilova2025multilingual} use the JSON format to detect propaganda techniques. To disambiguate spans with the same content, \citeauthor{klesnilova2025multilingual} proposes adding a JSON field \texttt{occurrence\_index} with the sequential number of the span.

\subsection{Issues with Existing Methods}
\label{sec:drawbacks}
As shown in Figure~\ref{fig:teaser}, all of the strategies are problematic. 
Some of the drawbacks are inherent, while others are related to specific limitations of LLMs.

\paragraph{LLMs cannot reliably \emph{copy}.} Both the \cattag{} and \catmatch{} strategies rely on the model's ability to copy its input. For \cattag{}, the model must reproduce the entire input text exactly to preserve alignment. For \catmatch{}, the generated span must match the input substring character-for-character. In practice, however, LLMs tend to ``fix'' the text by correcting typos, changing capitalization, or correcting factually incorrect sentences \cite{liu2024we,li2024prompting,jarolim2025can}. It is possible to partially mitigate this issue with fuzzy matching during post-processing, but that may introduce a new source of errors and does not solve the underlying issue.

\paragraph{LLMs cannot reliably \emph{count}.} Although positional embeddings allow LLMs to operate with relative positions of input tokens, the embeddings do not provide the model with a straightforward access to explicit token indices. The knowledge of number of characters in each token is learned indirectly \cite{edman2024cute,fu2024large}. Therefore, when \catindex{} strategies attempt to bypass text generation by predicting numerical positions directly, the indices are only ``informed guesses''. One can mitigate this by inserting indices into the text explicitly \cite{ramponi2025finegrained}. However, that disrupts the natural flow of the text and potentially degrades model performance.

\paragraph{Span content may be ambiguous.} For \catmatch{} strategies,  the lack of grounding in the input text can also introduce ambiguities. If the model correctly identifies a span (e.g., ``the''), but that exact span appears many times across the document, string matching alone cannot determine which specific occurrence the model intended to label. Disambiguating by asking the model to generate an occurrence index (where the model can suggest that it refers to the \emph{second} instance of word ``the''; \citealp{klesnilova2025multilingual}) once again places the burden on the model's counting abilities.

\section{\textsc{\ourmethod{}}: Making \emph{Matching} Methods More Robust}
\label{sec:our_method}

We introduce a new method -- \textsc{\ourmethod{}} -- that addresses the first weakness listed in \Cref{sec:drawbacks} for \catmatch{} strategies, i.e.,  the fact that the model may not perfectly reproduce the input span. Our method ensures that each decoded span is a valid input span while making only minimal interventions to the model's decoding process.

We implement \textsc{\ourmethod{}} for decoding a list of spans in a JSON format (cf.~\Cref{sec:json}). We ask the model to provide a list of JSON objects, where each object contains the field \texttt{text} with the span content.\footnote{The objects also contain other fields; we decode them without constraints. We only limit the \texttt{label} field we use for span categorization to contain valid labels.} The main idea of \textsc{\ourmethod{}} is to \textbf{limit the set of permissible tokens while the \texttt{text}\footnote{The field naming we selected is arbitrary and does not affect the generality of our method.} field is generated} to ensure that the decoded tokens are a continuous span from the input text.

Our method is orthogonal to the structured decoding approaches \cite{willard2023efficient}, which constrain the overall syntactical structure of the response, but cannot ensure that the \emph{content} of a specific field is a valid input substring. In fact, our approach is fully compatible with structured outputs; we combine the two methods in our experiments (cf. \Cref{sec:experiment_methods}).

\subsection{Algorithm}
Our algorithm (see \Cref{app:algmatch} for pseudocode) uses three modes: \texttt{DEFAULT}, \texttt{SELECT}, and \texttt{COPY}.

\paragraph{Default Mode} Our algorithm starts in \texttt{DEFAULT} mode, where it uses standard autoregressive decoding, sampling the next token from the model distribution: $y_t \sim P_{\mathcal{M}}(y_t \mid y_{<t}, X)$.

\paragraph{Select Mode} The \texttt{SELECT} mode is activated once the model starts decoding the value of the \texttt{text} field containing the span content. In \texttt{SELECT} mode, we limit the model's vocabulary to tokens present in the tokenized input $[x_1, x_2, \ldots, x_n]$ (here $x_i$ denotes the $i$-th input \emph{token}, not a character as in \Cref{sec:task_definition}) and let the model generate a \emph{single token} $x_i$ starting a span.

\paragraph{Copy Mode} After the token $x_i$ is generated, we switch to the \texttt{COPY} mode, where the model can generate the following input token $x_{i+k}$ (where initially $k=1$) or a closing quote (\texttt{"}). If the model generates the token $x_{i+k}$, we increment $k$ and continue in the \texttt{COPY} mode. In the latter case, we exit the \texttt{COPY} mode and continue in the \texttt{DEFAULT} mode. 

%The pseudocode for the algorithm is included in \Cref{app:algmatch}.

\subsection{Tokenization Issues}
\label{sec:tokenization}
Tokenization introduces additional aspects that need to be handled with care in \textsc{\ourmethod{}} to avoid disallowing valid tokens.

\paragraph{Different input and output tokenization.} We need to account for the fact that tokenization of the output text does \emph{not} necessarily correspond to the input text tokenization. Consider, for example, the input text \texttt{``Hello.''}. On the input, the text may be tokenized as \texttt{[``Hello'', ``.'']}. However, the model might first decode the token \texttt{``Hel''}, which guides it towards the tokens \texttt{``lo''} and \texttt{``.''}. Therefore, in the \texttt{COPY} mode, we need to allow not only the following input token, but \emph{all tokens that are valid prefixes of the subsequent text}.

\paragraph{Handling the field boundaries.} The quotes delimiting a JSON value may be part of a token containing other content-related characters. When entering the \texttt{SELECT} mode (i.e., in the step where the expression \texttt{``text''\textbackslash{}s:\textbackslash{}s``} would be decoded), we need to consider whether the token containing the opening quote also has a suffix (e.g. \texttt{``London}) and handle the suffix (in this case \texttt{London}) in the context of the \texttt{SELECT} mode. Accordingly, in the \texttt{COPY} mode, we need to whitelist all the tokens that satisfy the following: they start with the valid prefix, i.e., continuation of the decoded text, followed by a quotation mark, and optionally followed by a suffix (e.g., a comma followed by a newline); all these tokens exit the \texttt{COPY} mode.

\section{Experiments}
\label{sec:tasks}

\subsection{Methods}

\label{sec:experiment_methods}
Based on our survey in \Cref{sec:methods} (see also \Cref{tab:methods_overview}), we implement and evaluate what we consider a representative set of methods for each strategy: %We provide more details on individual methods, including model prompts and details on output parsing, in \Cref{app:prompts}. We implement the following methods:

\begin{itemize}
  \item \textsc{Tag}: \cattag{} with XML-like tags.
  \item \textsc{Index}: \catindex{} with character-level indices.
  \item \textsc{Index-Enriched}: \catindex{} with character-level indices, with the character indices inserted before each word on the input.
  \item \textsc{Match}: \catmatch{} with JSON output.
  \item \textsc{Match-Occ}: \catmatch{} with JSON output and occurrence index.
\end{itemize}

We also implement our novel method from \Cref{sec:our_method} analogically to the \catmatch{} methods:

\begin{itemize}
  \item \textsc{\ourmethod{}}: \catmatch{} with JSON output and constrained decoding.
  \item \textsc{\ourmethod{}-Occ}: \catmatch{} with JSON output, constrained decoding, and occurrence index.
\end{itemize}

For JSON-based outputs, we additionally test a structured output variant (denoted with \textsc{-S}) in which we enforce the JSON schema using structured output.\footnote{\url{https://docs.vllm.ai/en/stable/features/structured_outputs/}}
% \footnote{For \textsc{\ourmethod{}}, enforcing the JSON schema happens \emph{alongside} our custom constrained decoding. See \Cref{app:algmatch} for details.} 
In this variant, the model output is guaranteed to contain all the required fields and a valid category label. 

We parse the outputs of \cattag{} and \catindex{} approaches using regular expressions; \catmatch{} approaches use JSON parsing.
See \Cref{app:implementation_details} for more implementation details.

\subsection{Datasets}
For our experiments, we select four span labeling tasks. First, we use two well-established high-resource NLP tasks: named entity recognition and grammatical error correction.
% \footnote{Note that the inputs for NER and GEC use NLTK-style word tokenization \cite{bird2009natural} for separating punctuation from words, which we keep in its original form.} 
We also adopt span-level detection of errors in machine translation as a low-resource span labeling task. Finally, we create a synthetic task called \emph{conditional pattern lookup} to stress test robustness of all approaches. See \Cref{tab:datasets_overview} for an overview of our datasets.

\begin{table}[t]
\centering
\small
\begin{tabular}{lcccc}
\toprule
\textbf{Task} & \textbf{\# Ex.} & \textbf{\# Lang.} & \textbf{\# Cat.} & \textbf{Word tok.}\\
\midrule
NER & 7,523 & 13 & 3 & \ding{51}\\
GEC & 504 & 1 & 3 & \ding{51} \\
ESA-MT & 867 & 9 & 2 & \ding{55} \\
CPL & 1,000 & 1 & 1 & \ding{55} \\
\bottomrule
\end{tabular}
\caption{Statistics of the datasets used in our experiments. The table shows the number of examples, languages, and categories used for each task. The inputs for NER and GEC use NLTK-style word tokenization.}
\label{tab:datasets_overview}
\end{table}

\paragraph{Named Entity Recognition (NER)}
\label{sec:ner}

The goal in the NER task is to localize named entities and classify their type.
We use the UniversalNER benchmark \cite{mayhew2024universal} that covers three coarse-grained entity types: Person (\texttt{PER}), Organization (\texttt{ORG}), and Location (\texttt{LOC}). The dataset contains 7,523 examples across thirteen languages.

\paragraph{Grammatical Error Correction (GEC)}
\label{sec:gec}
In the GEC task, the goal is to identify spans requiring edits in order to make the text grammatically correct.
We adapt the English Write \& Improve corpus from the MultiGEC \cite{masciolini2025better} dataset that contains 504 examples with a list of edits between the original and corrected sentences from the ERRANT tool \cite{bryant-etal-2017-automatic}. We use the categorization of grammatical errors into three high-level categories: replacement (\texttt{R}), missing (\texttt{M}), and unnecessary (\texttt{U}).\footnote{For more details on how we adapted GEC for span labeling, see Appendix \ref{app:datasets}.}

\paragraph{Detecting Errors in Machine Translation (ESA-MT)}
\label{sec:mt} 

In the ESA-MT task, the goal is to find error spans in translated texts given their source.  We use data from the error span annotation (ESA) task at WMT24 \cite{kocmi2024error} that contains translation of texts from the news, social and literary domains from English into nine languages. Specifically, we select the examples containing errors from a test set of \citet{kasner2025large}, resulting in 867 examples. Following the dataset's annotation scheme, we classify the errors as \texttt{MINOR} or \texttt{MAJOR}.

\paragraph{Conditional Pattern Lookup (CPL)}
\label{sec:synth}
In real-world tasks, multiple occurrences of the same span are relatively rare. To test how the approaches behave with multiple occurrences of a specific span, we devise a synthetic task called \emph{conditional pattern lookup}.
The goal of the task is to find occurrences of a given regular expression in the input sequence. We use constraints to make the lookup conditional: only some occurrences of the pattern should be identified based on the pattern it is preceded or followed by. For example,  \textit{Find all sequences matching `\texttt{\textbackslash w+ dry}' that are not preceded by `\texttt{house}'}. Note that this can be easily solved by any regular expression matcher; we apply LLMs only for benchmarking purposes. As an input, we generate 1000 random sequences of approximately 100 English words. We ensure that the pattern occurs at least once in the input sequence. 
%We create two dataset splits: \textbf{CPL-o}, in which the spans may overlap, and \textbf{CPL-n} with no overlaps. In the following, \textbf{CPL} represents the union of these two splits.

\subsection{Models}

We test all approaches with state-of-the-art open LLMs: \includegraphics[height=1.5ex]{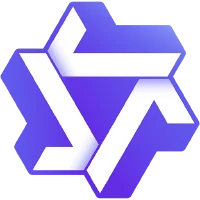}~\href{https://huggingface.co/Qwen/Qwen3-8B}{\text{Qwen3-8B}} \cite{yang2025qwen3}, 
  \includegraphics[height=1.5ex]{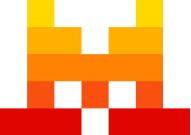}~\href{https://huggingface.co/mistralai/Mistral-Small-24B-Instruct-2501}{\text{Mistral-Small-24B-Instruct}} \cite{mistral2025small3}, 
  \includegraphics[height=1.5ex]{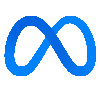}~\href{https://huggingface.co/meta-llama/Llama-3.3-70B-Instruct}{\text{Llama-3.3-70B-Instruct}} \cite{grattafiori2024llama}, and
  \includegraphics[height=1.5ex]{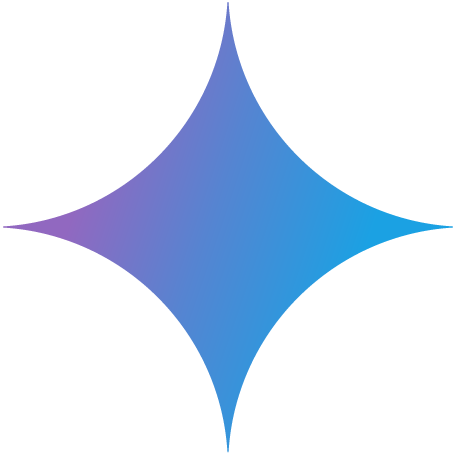}~\href{https://huggingface.co/google/gemma-4-31b-it}{\text{Gemma-4-31B-IT}} \cite{gemma4_2025}.
  In \Cref{app:gpt_results}, we provide additional results with the proprietary \includegraphics[height=1.5ex]{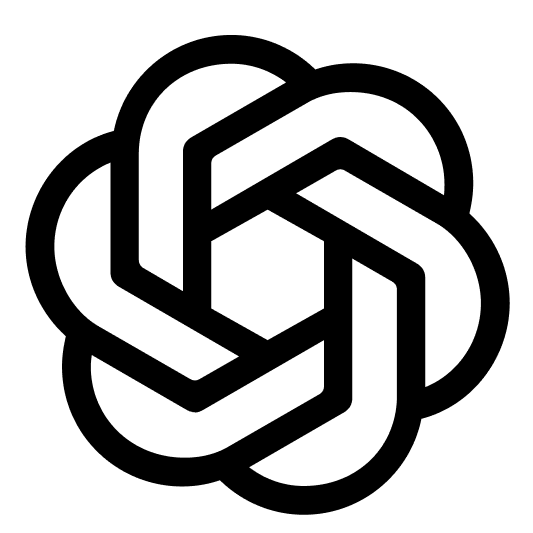}~\href{https://platform.openai.com/docs/models/gpt-5-mini}{GPT-5-mini} model for the methods that do not require logit access. 
  
  All models are run with 16-bit precision except for Llama, which is 4-bit quantized.
  We run each model with five different random seeds and average the results (see \Cref{app:setup} for details and standard deviations).  
  By default, we disable reasoning settings of the models. To test how reasoning may help, we performed additional experiments with Qwen3-8B with the reasoning effort set to \texttt{high}, which we denote as Qwen3-8B-Think; see \Cref{app:reasoning_experiments}.

\subsection{Evaluation Metric}
\label{sec:f1}
Our evaluation metric extends the standard F1-score by counting matches proportionally based on their character-level overlap \cite{dasanmartino2019finegrained,kasner2025large}. Let $C_p$ denote the set of character positions covered by predicted spans,
$C_g$ the set of character positions covered by gold spans, and $C_o = C_p \cap C_g$ the
positions covered by both. Then:
\begin{align*}
\text{Precision} = \frac{|C_o|}{|C_p|}, \quad
\text{Recall} = \frac{|C_o|}{|C_g|}, \\
\text{F1} = \frac{2 \cdot \text{Precision} \cdot \text{Recall}}{\text{Precision} + \text{Recall}}
\end{align*}

We evaluate both \emph{hard} and \emph{soft} variants of this metric. In the hard variant, a position in $C_o$ requires the gold and predicted spans covering it to share the same category label; the soft variant omits this requirement.

\begin{table*}[t]
  \centering
  % Auto-generated from analysis.ipynb
\small
\setlength{\tabcolsep}{5pt}
\renewcommand{\arraystretch}{1.08}
\begin{tabular*}{\textwidth}{l@{\extracolsep{12pt}}r@{\extracolsep{5pt}}r@{\extracolsep{5pt}}r@{\extracolsep{5pt}}r@{\extracolsep{12pt}}r@{\extracolsep{5pt}}r@{\extracolsep{5pt}}r@{\extracolsep{5pt}}r@{\extracolsep{12pt}}r@{\extracolsep{5pt}}r@{\extracolsep{5pt}}r@{\extracolsep{5pt}}r@{\extracolsep{12pt}}r@{\extracolsep{5pt}}r@{\extracolsep{5pt}}r@{\extracolsep{5pt}}r}
\toprule
 & \multicolumn{4}{c@{\extracolsep{\fill}}}{\textbf{NER}} & \multicolumn{4}{c@{\extracolsep{\fill}}}{\textbf{GEC}} & \multicolumn{4}{c@{\extracolsep{\fill}}}{\textbf{ESA-MT}} & \multicolumn{4}{c}{\textbf{CPL}} \\
\cmidrule(lr){2-5} \cmidrule(lr){6-9} \cmidrule(lr){10-13} \cmidrule(lr){14-17}
\textbf{Method} & \small \shortstack[c]{\includegraphics[height=1.5ex]{img/model_logos/qwen.png}\\ \texttt{8B}} & \small \shortstack[c]{\includegraphics[height=1.5ex]{img/model_logos/mistral.png}\\ \texttt{24B}} & \small \shortstack[c]{\includegraphics[height=1.5ex]{img/model_logos/llama.png}\\ \texttt{70B}} & \small \shortstack[c]{\includegraphics[height=1.5ex]{img/model_logos/gemma.png}\\ \texttt{31B}} & \small \shortstack[c]{\includegraphics[height=1.5ex]{img/model_logos/qwen.png}\\ \texttt{8B}} & \small \shortstack[c]{\includegraphics[height=1.5ex]{img/model_logos/mistral.png}\\ \texttt{24B}} & \small \shortstack[c]{\includegraphics[height=1.5ex]{img/model_logos/llama.png}\\ \texttt{70B}} & \small \shortstack[c]{\includegraphics[height=1.5ex]{img/model_logos/gemma.png}\\ \texttt{31B}} & \small \shortstack[c]{\includegraphics[height=1.5ex]{img/model_logos/qwen.png}\\ \texttt{8B}} & \small \shortstack[c]{\includegraphics[height=1.5ex]{img/model_logos/mistral.png}\\ \texttt{24B}} & \small \shortstack[c]{\includegraphics[height=1.5ex]{img/model_logos/llama.png}\\ \texttt{70B}} & \small \shortstack[c]{\includegraphics[height=1.5ex]{img/model_logos/gemma.png}\\ \texttt{31B}} & \small \shortstack[c]{\includegraphics[height=1.5ex]{img/model_logos/qwen.png}\\ \texttt{8B}} & \small \shortstack[c]{\includegraphics[height=1.5ex]{img/model_logos/mistral.png}\\ \texttt{24B}} & \small \shortstack[c]{\includegraphics[height=1.5ex]{img/model_logos/llama.png}\\ \texttt{70B}} & \small \shortstack[c]{\includegraphics[height=1.5ex]{img/model_logos/gemma.png}\\ \texttt{31B}} \\
\midrule
\textsc{Tag} & 73.3 & 74.1 & \textbf{82.8} & 88.9 & \textbf{27.2} & \textbf{31.3} & \textbf{34.7} & \textbf{52.9} & 10.2 & 11.0 & \textbf{10.1} & \textbf{14.2} & 40.2 & 53.1 & 79.5 & 93.3 \\
\addlinespace[3pt]
\hdashline[0.4pt/2pt]
\addlinespace[3pt]
\textsc{Index} & 19.1 & 27.1 & 27.7 & 53.1 & 10.4 & 7.2 & 8.5 & 9.4 & 7.5 & 7.4 & 6.1 & 7.9 & 34.1 & 40.0 & 36.6 & 57.9 \\
\textsc{Index-Enriched} & 33.7 & 68.5 & 70.0 & 87.4 & 11.0 & 18.5 & 16.4 & 48.3 & 7.5 & 8.1 & 9.0 & 9.8 & 45.4 & \textbf{70.1} & 84.4 & \textbf{98.6} \\
\addlinespace[3pt]
\hdashline[0.4pt/2pt]
\addlinespace[3pt]
\textsc{Match} & \textbf{74.1} & \textbf{79.1} & 81.7 & 87.4 & 12.1 & 15.5 & 20.1 & 38.4 & 11.4 & 12.2 & 9.2 & 12.4 & 42.8 & 43.8 & 48.5 & 51.1 \\
\textsc{Match-S} & 72.8 & 76.3 & 82.0 & 85.2 & 12.0 & 16.2 & 19.4 & 38.0 & 10.7 & \textbf{12.6} & 8.9 & 11.5 & 42.6 & 41.1 & 47.8 & 39.4 \\
\textsc{Match-Occ} & 73.0 & 78.3 & 81.6 & 88.2 & 15.4 & 18.6 & 20.8 & 38.6 & \textbf{11.9} & 12.4 & 9.4 & 13.0 & \textbf{70.1} & 69.1 & \textbf{86.5} & 94.7 \\
\textsc{Match-Occ-S} & 70.8 & 76.1 & 81.5 & 87.9 & 14.9 & 20.4 & 21.1 & 41.4 & 11.1 & 12.3 & 9.4 & 12.8 & \textbf{70.1} & 67.7 & 83.6 & 93.8 \\
\addlinespace[3pt]
\hdashline[0.4pt/2pt]
\addlinespace[3pt]
\textsc{\ourmethod{}} & 73.3 & 76.3 & 79.8 & 89.2 & 17.4 & 23.0 & 19.3 & 41.7 & 11.2 & 11.9 & 9.0 & 13.1 & 40.1 & 42.9 & 47.4 & 51.1 \\
\textsc{\ourmethod{}-S} & 71.9 & 73.8 & 79.9 & 86.2 & 16.8 & 20.4 & 21.6 & 41.7 & 10.8 & 11.4 & 8.8 & 11.4 & 41.2 & 40.9 & 46.7 & 39.7 \\
\textsc{\ourmethod{}-Occ} & 72.4 & 75.2 & 79.1 & \textbf{89.8} & 21.3 & 24.4 & 20.5 & 45.1 & 11.8 & 11.6 & 8.8 & 13.5 & 68.6 & 68.9 & 86.3 & 94.7 \\
\textsc{\ourmethod{}-Occ-S} & 70.1 & 73.0 & 79.3 & 89.2 & 20.4 & 23.5 & 22.7 & 45.2 & 11.2 & 11.1 & 9.0 & 12.8 & 69.6 & 68.0 & 84.4 & 93.7 \\
\bottomrule
\end{tabular*}

  \caption{Per-method \textbf{hard} F1 score in \%. The best score for each model per dataset is bold.}
  \label{tab:hardf1_by_method}
\end{table*}

% \begin{figure*}[t]
%     \centering
%     \includegraphics[width=\textwidth]{img/methods_bar_chart.pdf}
%     \caption{Error rates for each method (highlighted for each column separately), average for open LLMs across all datasets. We report three kinds of errors: error in parsing the model response, error in matching the span content, and error in matching the category label to one of the existing labels.}
%     \label{fig:methods}
% \end{figure*}

\begin{figure}[t]
    \centering
    \includegraphics[width=\columnwidth]{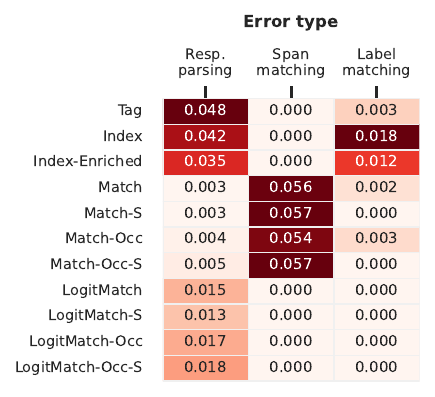}
    \caption{Error rates for each method (highlighted for each column separately), average for open LLMs across all datasets. We report three kinds of errors: error in parsing the model response, error in matching the span content, and error in matching the category label to one of the existing labels.}
    \label{fig:error_analysis}
\end{figure}

\section{Results}
\label{sec:experiments}

We primarily base our observations on the overlap-adjusted \emph{hard} F1-score in \Cref{tab:hardf1_by_method}. 
We also provide the \emph{soft} F1-score results in \Cref{tab:softf1_by_method} in \Cref{app:results}.
To give more insight into the failures of individual strategies, we also compute the following error rates (see \Cref{fig:error_analysis}):

\begin{itemize}
    \item \textbf{Response parsing errors}: The model output could not be parsed (e.g., invalid JSON).
    \item \textbf{Span matching errors}: The span content generated by the model could not be matched to any substring in the input text.
    \item \textbf{Label matching errors}: The category label generated by the model did not match any of the expected labels.
\end{itemize}

\begin{table*}[ht]
\centering
\small
\begin{tabular}{l p{13.5cm}}
\toprule
\multicolumn{2}{l}{\textbf{Task: NER} \quad \textbf{Model: Llama-3.3-70B}} \\
\midrule
Input & On Thursday , the 3rd of November , the mayor of \hlloc{Saint - Gaudens} ( in the \hlloc{Haute - Garonne} region ) decided to suspend the urban development agreement with the government formed in 2014 . \\
\addlinespace[1pt]
\hdashline[0.4pt/2pt]
\addlinespace[1pt]
Match & \texttt{\{"text": "\textcolor{red}{Saint-Gaudens}", "label": "LOC"\}},  \texttt{\{"text": "\textcolor{red}{Haute-Garonne}", "label": "LOC"\}} \\
\addlinespace[1pt]
\hdashline[0.4pt/2pt]
\addlinespace[1pt]
\ourmethod{} & \texttt{\{"text": "Saint - Gaudens", "label": "LOC"\}},  \texttt{\{"text": "Haute - Garonne", "label": "LOC"\}} \\
\midrule
\multicolumn{2}{l}{\textbf{Task: ESA-MT (en-cs)} \quad \textbf{Model: Gemma-4-31B-IT}} \\
\midrule
Source & \ldots{}what type of plane you'll be on, it's more complicated to determine what that means\ldots{} \\
Target (orig) & \ldots{}jaký typ letadla budete letět, ale složitější je určit, co to znamená\ldots{} \\
Target (enr.) & \ldots{}236::jaký  241::typ  245::letadla  253::budete  260::letět,  267::ale  271::složitější  282::je  285::určit\ldots{} \\
\addlinespace[1pt]
\hdashline[0.4pt/2pt]
\addlinespace[1pt]
Index & \texttt{\textcolor{red}{[247:261]} = MAJOR} \quad \textit{\textcolor{gray}{\ldots{}typ le\textcolor{red}{\underline{tadla budete l}}etět,\ldots{}}} \\
\addlinespace[2pt]
\hdashline[0.4pt/2pt]
\addlinespace[2pt]
Index-Enr. & \texttt{[253:266] = MAJOR} \quad \textit{\textcolor{gray}{\ldots{}letadla \underline{budete letět,} ale\ldots{}}} \\
\bottomrule
\end{tabular}
\caption{Qualitative examples comparing different span labeling methods. In the NER task, the \textbf{Match} method fails to reproduce the input tokenization (spaces around hyphens), while \textbf{\ourmethod{}} correctly adheres to it. In the ESA-MT task, the \textbf{Index} method generates indices that are ignoring word boundaries, whereas \textbf{Index-enriched} helps the model locate the span correctly. Problematic outputs are highlighted in \textcolor{red}{red}.}
\label{tab:qualitative_examples}
\end{table*}

\paragraph{\textsc{\ourmethod{}} improves matching performance in non-standard inputs.}
Our method improves the results on the NER and GEC tasks where the input text is tokenized using traditional NLP conventions (e.g., spaces around punctuation). With vanilla \catmatch{} methods, LLMs tend to normalize the text during generation, leading to mismatches (see \Cref{tab:qualitative_examples} for an example of such behavior). At the same time, our method maintains the performance of the \catmatch{} methods.

\paragraph{Tagging is robust, but token-heavy.} 

\begin{figure}[t]
\centering
\includegraphics[width=\columnwidth]{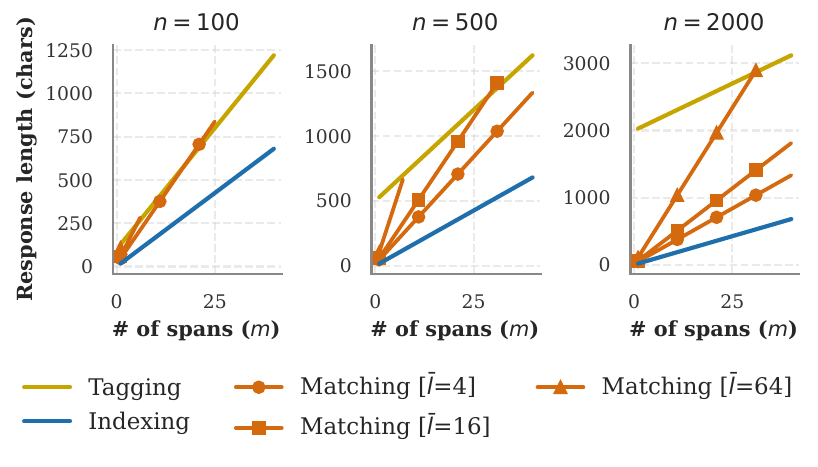}
\caption{Estimated response length of individual methods based on the length of the input text ($n$), the number of spans ($m$) and the average length of span content ($\bar{l}$).}
\label{fig:response_overhead}
\end{figure}

The \textsc{Tag} method has the most consistent performance across tasks. Notably, it outperforms other methods on GEC and shows competitive results for the other tasks. However, \catmatch{} methods are more token-efficient: \Cref{fig:response_overhead} shows the estimated response length of each method as a function of the input length ($n$), the span count ($m$), and the average span length ($\bar{l} = \frac{1}{m}\sum_i{e_i - b_i}$), derived from our specific output formats (see \Cref{app:token_efficiency} for details). As \cattag{} methods must reproduce the full input text in every response, their cost is usually the highest, unless the spans collectively cover most of the input text.

\paragraph{Indexing is not reliable without indices.}
Directly asking models to predict numerical indices (\textsc{Index}) leads to low performance on all tasks. As illustrated in \Cref{tab:qualitative_examples}, models frequently hallucinate indices that ignore word boundaries or fall entirely outside of the text. Adding explicit numbering to the input (\textsc{Index-Enriched}) can sometimes significantly improve the situation (e.g., by 24-52\% points on NER). However, this method still underperforms other methods, especially with weaker models. A plausible explanation is that the inserted indices interfere with the ability of the model to process the meaning of the input text.

% That being said, the tendencies to normalize the text do not seem to occur for the \cattag{} methods, where the model copies the input text end-to-end.

\paragraph{Using the \texttt{occurrence\_index} field helps with matching multiple occurrences.}

In the CPL task, the performance of \catmatch{} methods improves dramatically (by \textasciitilde 30-50\% points) by using the \texttt{occurrence\_index} for indexing the span content. This makes sense as \catmatch{} methods otherwise have no way to uniquely refer to individual occurrences of the span on the input. However, for tasks where the spans are mostly unique and unambiguous, such as NER and ESA-MT, this feature does not provide a clear benefit.

\paragraph{Structured outputs do not uniformly improve performance}
While structured outputs expectedly minimize parsing errors\footnote{The error rate is non-zero since the models may still fail to return a valid JSON, e.g., if they get stuck in an infinite loop and reach the maximum number of output tokens.} (see \Cref{fig:error_analysis}), restricting an output format naively may hurt model performance. Specifically, we find that the models \textit{without} structured outputs sometimes produce a ``spontaneous'' chain-of-thought before generating the output, which can help the model to improve its answer (at the cost of making the output longer).
% \footnote{We use regular expressions to extract the final output, see \Cref{app:implementation_details}.}

\paragraph{Explicit reasoning helps, but slows down inference.} The issue with restricting the output does not arise with reasoning models that explicitly enclose their reasoning trace between \texttt{<think>} tags, as this reasoning trace is handled separately in LLM frameworks.\footnote{Note that \textsc{\ourmethod{}} constrains only the final answer portion of the output (after the \texttt{</think>} closing tag); the reasoning trace itself is generated without any vocabulary constraints.} Our additional experiments (see \Cref{app:reasoning_experiments}) show that reasoning may boost the model performance -- for example, by explicitly indexing individual characters in the reasoning trace. However, this comes at the cost of a much higher number of output tokens during inference, which in our case sometimes lead to hitting the maximum output token limit before producing the answer.

\paragraph{Model size is not the only factor.} Since our primary goal was to explore span labeling strategies rather than benchmark models on the tasks, comparisons of model performance should be interpreted with caution. Our results suggest that Qwen3-8B is competitive with the much larger Llama-3.3-70B, and that Gemma-4-31B-IT achieves the highest scores overall on most tasks despite being a 31B model. Factors such as model release date, quantization, and training data may all influence these results.

\section{Conclusion}

We identified three major strategies for span labeling with LLMs and tested them on four span labeling tasks using state-of-the-art LLMs. Overall, the straightforward strategy of \cattag{} the spans with XML-like tags has the most consistent results. However, \catmatch{} strategies are more token-efficient and competitive overall. The \catindex{} strategies can be competitive only if the input is enriched with indices. We proposed a new method, \textsc{\ourmethod{}}, that solves the alignment issue of \catmatch{} methods.  Future work might explore how to combine the advantages of individual methods and how prompting with different examples and instructions influences the efficiency of each method.

\section*{Limitations}

We focused only on span labeling approaches that involve parsing output text, as we find these the most generally applicable. Therefore, we did not explore alternative approaches that are more resource intensive, such as LLM2Vec \cite{behnamghaderllm2vec} that transforms decoder-only LLMs into encoders with additional finetuning, or the approach of \citet{dukic2024looking} that allows the LLM to join span labeling with input processing by means of architecture modifications.

Our proposed method does not address the disambiguation of multiple identical spans. As we have shown, our method can be combined with \texttt{occurrence\_index} to achieve that effect. However, this approach is not perfect. Future work might thus explore other approaches, for example, disambiguating the span using attention values or by re-generating adjacent text.

In principle, a variant of \textsc{LogitMatch} is also applicable to \cattag{} methods, where it may constrain the decoded text to correspond to the input text. However, our preliminary experiments ran into implementation issues with tokenization handling, so we abandoned this approach. Future attempts should carefully handle tokens at the boundaries between the outer text, the tag itself, and the inner text to avoid performance degradation.

Finally, we note that NER and GEC are well-established tasks for which large amounts of BIO-tagged and XML-annotated training data are publicly available. Pre-training exposure to these annotation formats may advantage certain methods on those tasks. We selected ESA-MT and CPL as lower-resource tasks with less standard annotation formats to partially mitigate this effect, and we observe that the relative method rankings do indeed differ across tasks.

\section*{Ethical Considerations}

To the best of our knowledge, our work does not pose any immediate ethical risks. Our proposed method is designed to improve the reliability of text analysis tools, which we believe is a helpful contribution to the field. For our experiments, we relied solely on existing, publicly available datasets. We did not collect any new data or employ human annotators for this project. We used AI assistants to help with writing experimental code and to improve the clarity of the text. All generated content was manually reviewed and verified by the authors.

\section*{Acknowledgments}
This work was funded by the European Union (ERC, NG-NLG, 101039303), the
Charles University Research Centre program
No. 24/SSH/009, and the National Recovery Plan funded project MPO 60273/24/21300/21000 CEDMO 2.0 NPO. It used resources of the LINDAT/CLARIAH-CZ Research Infrastructure (Czech Ministry of Education, Youth, and Sports project No. LM2023062).

\bibliography{custom}

@inproceedings{behnamghaderllm2vec,
  title     = {LLM2Vec: Large Language Models Are Secretly Powerful Text Encoders},
  author    = {BehnamGhader, Parishad and Adlakha, Vaibhav and Mosbach, Marius and Bahdanau, Dzmitry and Chapados, Nicolas and Reddy, Siva},
  booktitle = {First Conference on Language Modeling},
  year      = {2024},
  url       = {https://openreview.net/forum?id=IW1PR7vEBf}
}

@inproceedings{dukic2024looking,
  title     = {Looking Right is Sometimes Right: Investigating the Capabilities of Decoder-only LLMs for Sequence Labeling},
  author    = {Duki{\'c}, David and {\v{S}}najder, Jan},
  booktitle = {Findings of the Association for Computational Linguistics ACL 2024},
  pages     = {14168--14181},
  year      = {2024},
  url       = {https://aclanthology.org/2024.findings-acl.843/}
}

@inproceedings{edman2024cute,
  address   = {USA},
  author    = {Lukas Edman and
               Helmut Schmid and
               Alexander Fraser},
  bibsource = {dblp computer science bibliography, https://dblp.org},
  booktitle = {Proceedings of the 2024 Conference on Empirical Methods in Natural Language Processing, {EMNLP} 2024, Miami, FL},
  doi       = {10.18653/V1/2024.EMNLP-MAIN.177},
  editor    = {Yaser Al{-}Onaizan and
               Mohit Bansal and
               Yun{-}Nung Chen},
  pages     = {3017--3026},
  title     = {{CUTE:} {M}easuring {L}{L}{M}s' {U}nderstanding of {T}heir {T}okens},
  url       = {https://doi.org/10.18653/v1/2024.emnlp-main.177},
  year      = {2024}
}

@article{fu2024large,
  author     = {Tairan Fu and
                Raquel Ferrando and
                Javier Conde and
                Carlos Arriaga and
                Pedro Reviriego},
  bibsource  = {dblp computer science bibliography, https://dblp.org},
  doi        = {10.48550/ARXIV.2412.18626},
  eprint     = {2412.18626},
  eprinttype = {arXiv},
  journal    = {CoRR},
  title      = {{W}hy {D}o {L}arge {L}anguage {M}odels ({L}{L}{M}s) {S}truggle to {C}ount {L}etters?},
  url        = {https://doi.org/10.48550/arXiv.2412.18626},
  volume     = {abs/2412.18626},
  year       = {2024}
}

@inproceedings{kwon2023efficient,
  address   = {Germany},
  author    = {Woosuk Kwon and
               Zhuohan Li and
               Siyuan Zhuang and
               Ying Sheng and
               Lianmin Zheng and
               Cody Hao Yu and
               Joseph Gonzalez and
               Hao Zhang and
               Ion Stoica},
  bibsource = {dblp computer science bibliography, https://dblp.org},
  booktitle = {Proceedings of the 29th Symposium on Operating Systems Principles, {SOSP} 2023, Koblenz},
  doi       = {10.1145/3600006.3613165},
  editor    = {Jason Flinn and
               Margo I. Seltzer and
               Peter Druschel and
               Antoine Kaufmann and
               Jonathan Mace},
  pages     = {611--626},
  title     = {{E}fficient {M}emory {M}anagement for {L}arge {L}anguage {M}odel {S}erving with {P}aged{A}ttention},
  url       = {https://doi.org/10.1145/3600006.3613165},
  year      = {2023}
}

@misc{mistral2025small3,
  author       = {{Mistral AI Team}},
  day          = {30},
  howpublished = {\url{https://mistral.ai/news/mistral-small-3}},
  month        = jan,
  note         = {Accessed: 2026-01-02},
  title        = {{M}istral {S}mall 3},
  year         = {2025}
}

@article{grattafiori2024llama,
  author  = {Grattafiori, Aaron and Dubey, Abhimanyu and Jauhri, Abhinav and Pandey, Abhinav and Kadian, Abhishek and Al-Dahle, Ahmad and Letman, Aiesha and Mathur, Akhil and Schelten, Alan and Vaughan, Alex and others},
  journal = {arXiv preprint arXiv:2407.21783},
  title   = {{T}he llama 3 herd of models},
  year    = {2024},
  url     = {https://arxiv.org/abs/2407.21783}
}

@article{yang2025qwen3,
  author     = {An Yang and
                Anfeng Li and
                Baosong Yang and
                Beichen Zhang and
                Binyuan Hui and
                Bo Zheng and
                Bowen Yu and
                Chang Gao and
                Chengen Huang and
                Chenxu Lv and
                Chujie Zheng and
                Dayiheng Liu and
                Fan Zhou and
                Fei Huang and
                Feng Hu and
                Hao Ge and
                Haoran Wei and
                Huan Lin and
                Jialong Tang and
                Jian Yang and
                Jianhong Tu and
                Jianwei Zhang and
                Jian Yang and
                Jiaxi Yang and
                Jingren Zhou and
                Junyang Lin and
                Kai Dang and
                Keqin Bao and
                Kexin Yang and
                Le Yu and
                Lianghao Deng and
                Mei Li and
                Mingfeng Xue and
                Mingze Li and
                Pei Zhang and
                Peng Wang and
                Qin Zhu and
                Rui Men and
                Ruize Gao and
                Shixuan Liu and
                Shuang Luo and
                Tianhao Li and
                Tianyi Tang and
                Wenbiao Yin and
                Xingzhang Ren and
                Xinyu Wang and
                Xinyu Zhang and
                Xuancheng Ren and
                Yang Fan and
                Yang Su and
                Yichang Zhang and
                Yinger Zhang and
                Yu Wan and
                Yuqiong Liu and
                Zekun Wang and
                Zeyu Cui and
                Zhenru Zhang and
                Zhipeng Zhou and
                Zihan Qiu},
  bibsource  = {dblp computer science bibliography, https://dblp.org},
  doi        = {10.48550/ARXIV.2505.09388},
  eprint     = {2505.09388},
  eprinttype = {arXiv},
  journal    = {CoRR},
  title      = {{Q}wen3 {T}echnical {R}eport},
  url        = {https://doi.org/10.48550/arXiv.2505.09388},
  volume     = {abs/2505.09388},
  year       = {2025}
}

@inproceedings{li2024prompting,
  address   = {Torino, Italy},
  author    = {Yongqi Li and
               Mayi Xu and
               Xin Miao and
               Shen Zhou and
               Tieyun Qian},
  bibsource = {dblp computer science bibliography, https://dblp.org},
  booktitle = {Proceedings of the 2024 Joint International Conference on Computational Linguistics, Language Resources and Evaluation, {LREC/COLING} 2024, 20-25 May, 2024},
  editor    = {Nicoletta Calzolari and
               Min{-}Yen Kan and
               V{\'{e}}ronique Hoste and
               Alessandro Lenci and
               Sakriani Sakti and
               Nianwen Xue},
  pages     = {13201--13221},
  title     = {{P}rompting {L}arge {L}anguage {M}odels for {C}ounterfactual {G}eneration: {A}n {E}mpirical {S}tudy},
  url       = {https://aclanthology.org/2024.lrec-main.1156},
  year      = {2024}
}

@inproceedings{liu2024we,
  address   = {Honolulu, HI, USA},
  author    = {Michael Xieyang Liu and
               Frederick Liu and
               Alexander J. Fiannaca and
               Terry Koo and
               Lucas Dixon and
               Michael Terry and
               Carrie J. Cai},
  bibsource = {dblp computer science bibliography, https://dblp.org},
  booktitle = {Extended Abstracts of the {CHI} Conference on Human Factors in Computing Systems, {CHI} {EA} 2024},
  doi       = {10.1145/3613905.3650756},
  editor    = {Florian 'Floyd' Mueller and
               Penny Kyburz and
               Julie R. Williamson and
               Corina Sas},
  pages     = {10:1--10:9},
  title     = {"{W}e {N}eed {S}tructured {O}utput": {T}owards {U}ser-centered {C}onstraints on {L}arge {L}anguage {M}odel {O}utput},
  url       = {https://doi.org/10.1145/3613905.3650756},
  year      = {2024}
}

@article{jarolim2025can,
  author  = {Jarol{\'\i}m, Anton{\'\i}n and Faj{\v{c}}{\'\i}k, Martin and Makaiov{\'a}, Lucia},
  journal = {arXiv preprint arXiv:2511.21401},
  title   = {{C}an {L}{L}{M}s extract human-like fine-grained evidence for evidence-based fact-checking?},
  year    = {2025},
  url     = {https://arxiv.org/abs/2511.21401}
}

@misc{guidance,
  author       = {{guidance-ai}},
  howpublished = {\url{https://github.com/guidance-ai/guidance}},
  note         = {MIT License},
  title        = {{G}uidance: {A} language for constraining large language models},
  year         = {2025}
}

@article{willard2023efficient,
  author     = {Brandon T. Willard and
                R{\'{e}}mi Louf},
  bibsource  = {dblp computer science bibliography, https://dblp.org},
  doi        = {10.48550/ARXIV.2307.09702},
  eprint     = {2307.09702},
  eprinttype = {arXiv},
  journal    = {CoRR},
  title      = {{E}fficient {G}uided {G}eneration for {L}arge {L}anguage {M}odels},
  url        = {https://doi.org/10.48550/arXiv.2307.09702},
  volume     = {abs/2307.09702},
  year       = {2023}
}

@inproceedings{vaswani2017attention,
  author    = {Ashish Vaswani and
               Noam Shazeer and
               Niki Parmar and
               Jakob Uszkoreit and
               Llion Jones and
               Aidan N. Gomez and
               Lukasz Kaiser and
               Illia Polosukhin},
  bibsource = {dblp computer science bibliography, https://dblp.org},
  booktitle = {Advances in Neural Information Processing Systems 30: Annual Conference
               on Neural Information Processing Systems 2017, December 4-9, 2017,
               Long Beach, CA, {USA}},
  editor    = {Isabelle Guyon and
               Ulrike von Luxburg and
               Samy Bengio and
               Hanna M. Wallach and
               Rob Fergus and
               S. V. N. Vishwanathan and
               Roman Garnett},
  pages     = {5998--6008},
  title     = {{A}ttention is {A}ll you {N}eed},
  url       = {https://proceedings.neurips.cc/paper/2017/hash/3f5ee243547dee91fbd053c1c4a845aa-Abstract.html},
  year      = {2017}
}

@article{radford2019language,
  author  = {Radford, Alec and Wu, Jeffrey and Child, Rewon and Luan, David and Amodei, Dario and Sutskever, Ilya and others},
  journal = {OpenAI blog},
  number  = {8},
  pages   = {9},
  title   = {{L}anguage models are unsupervised multitask learners},
  volume  = {1},
  year    = {2019},
  url     = {https://cdn.openai.com/better-language-models/language_models_are_unsupervised_multitask_learners.pdf}
}

@inproceedings{ramshaw-marcus-1995-text,
  address   = {Cambridge, Massachusetts, USA},
  author    = {Lance A. Ramshaw and
               Mitch Marcus},
  bibsource = {dblp computer science bibliography, https://dblp.org},
  booktitle = {Third Workshop on Very Large Corpora, VLC at ACL 1995},
  editor    = {David Yarowsky and
               Kenneth Church},
  title     = {{T}ext {C}hunking using {T}ransformation-Based {L}earning},
  url       = {https://aclanthology.org/W95-0107/},
  year      = {1995}
}

@inproceedings{dasanmartino2019finegrained,
  author     = {Da San Martino, Giovanni and Yu, Seunghak and Barrón-Cedeño, Alberto and Petrov, Rostislav and Nakov, Preslav},
  booktitle  = {Proceedings of the 2019 {{Conference}} on {{Empirical Methods}} in {{Natural Language Processing}} and the 9th {{International Joint Conference}} on {{Natural Language Processing}} ({{EMNLP-IJCNLP}})},
  doi        = {10.18653/v1/D19-1565},
  editor     = {Inui, Kentaro and Jiang, Jing and Ng, Vincent and Wan, Xiaojun},
  eventtitle = {{{EMNLP-IJCNLP}} 2019},
  location   = {Hong Kong, China},
  pages      = {5636--5646},
  publisher  = {Association for Computational Linguistics},
  title      = {Fine-{{Grained Analysis}} of {{Propaganda}} in {{News Article}}},
  url        = {https://aclanthology.org/D19-1565},
  year       = {2019}
}

@inproceedings{wang2023gptnernamedentityrecognition,
  address   = {Albuquerque, New Mexico, USA},
  author    = {Shuhe Wang and
               Xiaofei Sun and
               Xiaoya Li and
               Rongbin Ouyang and
               Fei Wu and
               Tianwei Zhang and
               Jiwei Li and
               Guoyin Wang and
               Chen Guo},
  bibsource = {dblp computer science bibliography, https://dblp.org},
  booktitle = {Findings of the Association for Computational Linguistics: {NAACL} 2025},
  doi       = {10.18653/V1/2025.FINDINGS-NAACL.239},
  editor    = {Luis Chiruzzo and
               Alan Ritter and
               Lu Wang},
  pages     = {4257--4275},
  title     = {{GPT-NER:} {N}amed {E}ntity {R}ecognition via {L}arge {L}anguage {M}odels},
  url       = {https://doi.org/10.18653/v1/2025.findings-naacl.239},
  year      = {2025}
}

@inproceedings{warner2024smarterbetterfasterlonger,
  address   = {Vienna, Austria},
  author    = {Benjamin Warner and
               Antoine Chaffin and
               Benjamin Clavi{\'{e}} and
               Orion Weller and
               Oskar Hallstr{\"{o}}m and
               Said Taghadouini and
               Alexis Gallagher and
               Raja Biswas and
               Faisal Ladhak and
               Tom Aarsen and
               Griffin Thomas Adams and
               Jeremy Howard and
               Iacopo Poli},
  bibsource = {dblp computer science bibliography, https://dblp.org},
  booktitle = {Proceedings of the 63rd Annual Meeting of the Association for Computational Linguistics (Volume 1: Long Papers), {ACL} 2025},
  editor    = {Wanxiang Che and
               Joyce Nabende and
               Ekaterina Shutova and
               Mohammad Taher Pilehvar},
  pages     = {2526--2547},
  title     = {{S}marter, {B}etter, {F}aster, {L}onger: {A} {M}odern {B}idirectional {E}ncoder for {F}ast, {M}emory {E}fficient, and {L}ong {C}ontext {F}inetuning and {I}nference},
  url       = {https://aclanthology.org/2025.acl-long.127/},
  year      = {2025}
}

@inproceedings{devlin2019bertpretrainingdeepbidirectional,
  address   = {Minneapolis, MN, USA},
  author    = {Jacob Devlin and
               Ming{-}Wei Chang and
               Kenton Lee and
               Kristina Toutanova},
  bibsource = {dblp computer science bibliography, https://dblp.org},
  booktitle = {Proceedings of the 2019 Conference of the North American Chapter of the Association for Computational Linguistics: Human Language Technologies, {NAACL-HLT} 2019, Volume 1 (Long and Short Papers)},
  doi       = {10.18653/V1/N19-1423},
  editor    = {Jill Burstein and
               Christy Doran and
               Thamar Solorio},
  pages     = {4171--4186},
  title     = {{BERT:} {P}re-training of {D}eep {B}idirectional {T}ransformers for {L}anguage {U}nderstanding},
  url       = {https://doi.org/10.18653/v1/n19-1423},
  year      = {2019}
}

@inproceedings{bryant-etal-2017-automatic,
  address   = {Vancouver, Canada},
  author    = {Christopher Bryant and
               Mariano Felice and
               Ted Briscoe},
  bibsource = {dblp computer science bibliography, https://dblp.org},
  booktitle = {Proceedings of the 55th Annual Meeting of the Association for Computational Linguistics, {ACL} 2017 4, Volume 1: Long Papers},
  doi       = {10.18653/V1/P17-1074},
  editor    = {Regina Barzilay and
               Min{-}Yen Kan},
  pages     = {793--805},
  title     = {{A}utomatic {A}nnotation and {E}valuation of {E}rror {T}ypes for {G}rammatical {E}rror {C}orrection},
  url       = {https://doi.org/10.18653/v1/P17-1074},
  year      = {2017}
}

@inproceedings{ramponi2025finegrained,
  address   = {New Mexico, USA},
  author    = {Alan Ramponi and
               Agnese Daffara and
               Sara Tonelli},
  bibsource = {dblp computer science bibliography, https://dblp.org},
  booktitle = {Proceedings of the 2025 Conference of the Nations of the Americas Chapter of the Association for Computational Linguistics: Human Language Technologies, {NAACL} 2025 - Volume 1: Long Papers, Albuquerque},
  doi       = {10.18653/V1/2025.NAACL-LONG.34},
  editor    = {Luis Chiruzzo and
               Alan Ritter and
               Lu Wang},
  pages     = {762--784},
  title     = {{F}ine-grained {F}allacy {D}etection with {H}uman {L}abel {V}ariation},
  url       = {https://doi.org/10.18653/v1/2025.naacl-long.34},
  year      = {2025}
}

@inproceedings{hasanain2024large,
  address   = {Miami, Florida, USA},
  author    = {Maram Hasanain and
               Fatema Ahmad and
               Firoj Alam},
  bibsource = {dblp computer science bibliography, https://dblp.org},
  booktitle = {Findings of the Association for Computational Linguistics: {EMNLP} 2024},
  doi       = {10.18653/V1/2024.FINDINGS-EMNLP.850},
  editor    = {Yaser Al{-}Onaizan and
               Mohit Bansal and
               Yun{-}Nung Chen},
  pages     = {14522--14532},
  title     = {{L}arge {L}anguage {M}odels for {P}ropaganda {S}pan {A}nnotation},
  url       = {https://doi.org/10.18653/v1/2024.findings-emnlp.850},
  year      = {2024}
}

@article{klesnilova2025multilingual,
  author = {Klesnilová, Kristýna},
  title  = {{M}ultilingual {D}etection of {P}ersuasion {T}echniques in {T}ext using {L}arge {L}anguage {M}odels},
  year   = {2025},
  url    = {https://dspace.cuni.cz/handle/20.500.11956/203238}
}

@inproceedings{kasner2024traditional,
  address   = {Bangkok, Thailand},
  author    = {Zdeněk Kasner and
               Ondřej Dušek},
  bibsource = {dblp computer science bibliography, https://dblp.org},
  booktitle = {Proceedings of the 62nd Annual Meeting of the Association for Computational Linguistics (Volume 1: Long Papers), {ACL} 2024},
  doi       = {10.18653/V1/2024.ACL-LONG.651},
  editor    = {Lun{-}Wei Ku and
               Andre Martins and
               Vivek Srikumar},
  pages     = {12045--12072},
  title     = {{B}eyond {T}raditional {B}enchmarks: {A}nalyzing {B}ehaviors of {O}pen {L}{L}{M}s on {D}ata-to-Text {G}eneration},
  url       = {https://doi.org/10.18653/v1/2024.acl-long.651},
  year      = {2024}
}

@misc{kasner2025large,
  title         = {LLMs as Span Annotators: A Comparative Study of LLMs and Humans},
  author        = {Zdeněk Kasner and Vilém Zouhar and Patrícia Schmidtová and Ivan Kartáč and Kristýna Onderková and Ondřej Plátek and Dimitra Gkatzia and Saad Mahamood and Ondřej Dušek and Simone Balloccu},
  year          = {2025},
  eprint        = {2504.08697},
  archiveprefix = {arXiv},
  primaryclass  = {cs.CL},
  url           = {https://arxiv.org/abs/2504.08697}
}

@inproceedings{kocmi2023gembamqm,
  address   = {Singapore},
  author    = {Tom Kocmi and
               Christian Federmann},
  bibsource = {dblp computer science bibliography, https://dblp.org},
  booktitle = {Proceedings of the Eighth Conference on Machine Translation, {WMT} 2023},
  doi       = {10.18653/V1/2023.WMT-1.64},
  editor    = {Philipp Koehn and
               Barry Haddon and
               Tom Kocmi and
               Christof Monz},
  pages     = {768--775},
  title     = {{GEMBA-MQM:} {D}etecting {T}ranslation {Q}uality {E}rror {S}pans with {GPT-4}},
  url       = {https://doi.org/10.18653/v1/2023.wmt-1.64},
  year      = {2023}
}

@inproceedings{obeidat2025llms,
  address   = {Rende, Italy},
  author    = {Motasem S. Obeidat and
               Md Sultan Al Nahian and
               Ramakanth Kavuluru},
  bibsource = {dblp computer science bibliography, https://dblp.org},
  booktitle = {13th {IEEE} International Conference on Healthcare Informatics, {ICHI} 2025},
  doi       = {10.1109/ICHI64645.2025.00048},
  pages     = {352--358},
  title     = {{D}o {L}{L}{M}s {S}urpass {E}ncoders for {B}iomedical {N}{E}{R}?},
  url       = {https://doi.org/10.1109/ICHI64645.2025.00048},
  year      = {2025}
}

@inproceedings{treviso2024xtower,
  address   = {Miami, Florida, USA},
  author    = {Marcos V. Treviso and
               Nuno Miguel Guerreiro and
               Sweta Agrawal and
               Ricardo Rei and
               Jos{\'{e}} Pombal and
               T{\^{a}}nia Vaz and
               Helena Wu and
               Beatriz Silva and
               Daan van Stigt and
               Andr{\'{e}} F. T. Martins},
  bibsource = {dblp computer science bibliography, https://dblp.org},
  booktitle = {Findings of the Association for Computational Linguistics: {EMNLP} 2024},
  doi       = {10.18653/V1/2024.FINDINGS-EMNLP.892},
  editor    = {Yaser Al{-}Onaizan and
               Mohit Bansal and
               Yun{-}Nung Chen},
  pages     = {15222--15239},
  title     = {x{T}ower: {A} {M}ultilingual {LLM} for {E}xplaining and {C}orrecting {T}ranslation {E}rrors},
  url       = {https://doi.org/10.18653/v1/2024.findings-emnlp.892},
  year      = {2024}
}

@inproceedings{yan2024ltner,
  address   = {Kolkata, India},
  author    = {Faren Yan and
               Peng Yu and
               Xin Chen},
  bibsource = {dblp computer science bibliography, https://dblp.org},
  booktitle = {Pattern Recognition - 27th International Conference, {ICPR} 2024},
  doi       = {10.1007/978-3-031-78495-8\_25},
  editor    = {Apostolos Antonacopoulos and
               Subhasis Chaudhuri and
               Rama Chellappa and
               Cheng{-}Lin Liu and
               Saumik Bhattacharya and
               Umapada Pal},
  pages     = {399--411},
  series    = {Lecture Notes in Computer Science},
  title     = {{LTNER:} {L}arge {L}anguage {M}odel {T}agging for {N}amed {E}ntity {R}ecognition with {C}ontextualized {E}ntity {M}arking},
  url       = {https://doi.org/10.1007/978-3-031-78495-8\_25},
  volume    = {15319},
  year      = {2024}
}

@inproceedings{wang2023gptner,
  address   = {Albuquerque, New Mexico, USA},
  author    = {Shuhe Wang and
               Xiaofei Sun and
               Xiaoya Li and
               Rongbin Ouyang and
               Fei Wu and
               Tianwei Zhang and
               Jiwei Li and
               Guoyin Wang and
               Chen Guo},
  bibsource = {dblp computer science bibliography, https://dblp.org},
  booktitle = {Findings of the Association for Computational Linguistics: {NAACL} 2025},
  doi       = {10.18653/V1/2025.FINDINGS-NAACL.239},
  editor    = {Luis Chiruzzo and
               Alan Ritter and
               Lu Wang},
  pages     = {4257--4275},
  title     = {{GPT-NER:} {N}amed {E}ntity {R}ecognition via {L}arge {L}anguage {M}odels},
  url       = {https://doi.org/10.18653/v1/2025.findings-naacl.239},
  year      = {2025}
}

@inproceedings{kocmi2024error,
  address   = {USA},
  author    = {Tom Kocmi and
               Vil{\'{e}}m Zouhar and
               Eleftherios Avramidis and
               Roman Grundkiewicz and
               Marzena Karpinska and
               Maja Popovic and
               Mrinmaya Sachan and
               Mariya Shmatova},
  bibsource = {dblp computer science bibliography, https://dblp.org},
  booktitle = {Proceedings of the Ninth Conference on Machine Translation, {WMT} 2024, Miami, FL},
  doi       = {10.18653/V1/2024.WMT-1.131},
  editor    = {Barry Haddow and
               Tom Kocmi and
               Philipp Koehn and
               Christof Monz},
  pages     = {1440--1453},
  title     = {{E}rror {S}pan {A}nnotation: {A} {B}alanced {A}pproach for {H}uman {E}valuation of {M}achine {T}ranslation},
  url       = {https://doi.org/10.18653/v1/2024.wmt-1.131},
  year      = {2024}
}

@article{masciolini2025better,
  author     = {Masciolini, Arianna and Caines, Andrew and De Clercq, Orphée and Kruijsbergen, Joni and Kurfal, Murathan and Muñoz Sánchez, Ricardo and Volodina, Elena and Östling, Robert and Allkivi, Kais and Arhar Holdt, Špela and Auzina, Ilze and Darģis, Roberts and Drakonaki, Elena and Frey, Jennifer-Carmen and Glišić, Isidora and Kikilintza, Pinelopi and Nicolas, Lionel and Romanyshyn, Mariana and Rosen, Alexandr and Rozovskaya, Alla and Suluste, Kristjan and Syvokon, Oleksiy and Tantos, Alexandros and Touriki, Despoina-Ourania and Tsiotskas, Konstantinos and Tsourilla, Eleni and Varsamopoulos, Vassilis and Wisniewski, Katrin and Žagar, Aleš and Zesch, Torsten},
  copyright  = {http://creativecommons.org/licenses/by/4.0/},
  doi        = {10.1075/ijlcr.24033.mas},
  issn       = {2215-1478, 2215-1486},
  journal    = {International Journal of Learner Corpus Research},
  langid     = {english},
  pages      = {309--335},
  shorttitle = {Towards Better Language Representation in {{Natural Language Processing}}},
  title      = {{T}owards {B}etter {L}anguage {R}epresentation in {{Natural {L}anguage {P}rocessing}}: {{A}} {M}ultilingual {D}ataset for {T}ext-Level {{Grammatical {E}rror {C}orrection}}},
  url        = {http://www.jbe-platform.com/content/journals/10.1075/ijlcr.24033.mas},
  volume     = {11},
  year       = {2025}
}

@inproceedings{mayhew2024universal,
  address   = {Mexico City, Mexico},
  author    = {Stephen Mayhew and
               Terra Blevins and
               Shuheng Liu and
               Marek Suppa and
               Hila Gonen and
               Joseph Marvin Imperial and
               B{\"{o}}rje Karlsson and
               Peiqin Lin and
               Nikola Ljubesic and
               Lester James V. Miranda and
               Barbara Plank and
               Arij Riabi and
               Yuval Pinter},
  bibsource = {dblp computer science bibliography, https://dblp.org},
  booktitle = {Proceedings of the 2024 Conference of the North American Chapter of the Association for Computational Linguistics: Human Language Technologies (Volume 1: Long Papers), {NAACL} 2024},
  doi       = {10.18653/V1/2024.NAACL-LONG.243},
  editor    = {Kevin Duh and
               Helena G{\'{o}}mez{-}Adorno and
               Steven Bethard},
  pages     = {4322--4337},
  title     = {{U}niversal {NER:} {A} {G}old-Standard {M}ultilingual {N}amed {E}ntity {R}ecognition {B}enchmark},
  url       = {https://doi.org/10.18653/v1/2024.naacl-long.243},
  year      = {2024}
}

@inproceedings{mulcaire2025span,
  address    = {Vienna, Austria},
  author     = {Mulcaire, Phoebe and Madnani, Nitin},
  booktitle  = {Proceedings of the 20th {{Workshop}} on {{Innovative Use}} of {{NLP}} for {{Building Educational Applications}} ({{BEA}} 2025)},
  doi        = {10.18653/v1/2025.bea-1.62},
  editor     = {Kochmar, Ekaterina and Alhafni, Bashar and Bexte, Marie and Burstein, Jill and Horbach, Andrea and {Laarmann-Quante}, Ronja and Tack, Anaïs and Yaneva, Victoria and Yuan, Zheng},
  isbn       = {979-8-89176-270-1},
  pages      = {850--859},
  shorttitle = {Span {{Labeling}} with {{Large Language Models}}},
  title      = {{S}pan {{Labeling}} with {{Large {L}anguage {M}odels}}: {{Shell}} vs. {{Meat}}},
  url        = {https://aclanthology.org/2025.bea-1.62/},
  year       = {2025}
}

@article{thomson2023evaluating,
  author    = {Craig Thomson and
               Ehud Reiter and
               Barkavi Sundararajan},
  bibsource = {dblp computer science bibliography, https://dblp.org},
  doi       = {10.1016/J.CSL.2023.101482},
  journal   = {Comput. Speech Lang.},
  pages     = {101482},
  title     = {{E}valuating factual accuracy in complex data-to-text},
  url       = {https://doi.org/10.1016/j.csl.2023.101482},
  volume    = {80},
  year      = {2023}
}

@misc{gemma4_2025,
  title  = {Gemma 4},
  author = {{Google DeepMind}},
  year   = {2025},
  url    = {https://deepmind.google/models/gemma/gemma-4/},
  note   = {Apache 2.0 license}
}

\appendix

\crefalias{section}{appendix}
\crefalias{subsection}{appendix}

% \section{Implementation Details}
% \label{app:implementation_details}

% In this section, we describe additional details regarding our data processing (§\ref{app:datasets}), implementation (§\ref{app:prompts}), and experimental setup (§\ref{app:setup}).

\section{Experimental Setup}
\label{app:setup}
\subsection{Models} 
\Cref{tab:model_params} presents the overview of our models and their decoding parameters. We run the open models using vLLM \cite{kwon2023efficient} in 16-bit precision, except for Llama 3.3 70B that we quantize to 4-bit precision. For structured outputs, we use the vLLM structured output feature.\footnote{\url{https://docs.vllm.ai/en/latest/features/structured_outputs/}} The decoding parameters are the default parameters in the Huggingface configuration file. We set only the \texttt{max\_tokens} parameter manually to 4,096 (and increased it to 16,384 for the reasoning models, see \Cref{app:reasoning_experiments}). 

\subsection{Multiple runs} 
For each experiment, we ran each configuration with 5 different random seeds (42--46) and report the mean across seeds. Averaging over method-model configurations, the mean cross-seed standard deviation of hard F1 in percentage points is 0.46 on NER, 0.96 on GEC, 0.51 on ESA-MT, and 0.98 on CPL (see \Cref{tab:hardf1_seed_std}). These values indicate that the overall ranking patterns are stable across random seeds.

\begin{table*}[ht]
\centering
\small
\begin{tabular}{lp{8cm}cccc}
\toprule
\textbf{Model} & \textbf{Model ID} & \textbf{temp.} & \textbf{top-p} & \textbf{top-k}  \\
\midrule
\includegraphics[height=1.5ex]{img/model_logos/qwen.png} Qwen3-8B & \texttt{Qwen/Qwen3-8B} & 0.6 & 0.95 & 20 \\
\includegraphics[height=1ex]{img/model_logos/mistral.png} Mistral-Small-24B-Instruct & \texttt{mistralai/Mistral-Small-24B-Instruct-2501} & 0.15 & 1.0 & - \\
\includegraphics[height=1.5ex]{img/model_logos/llama.png} Llama-3.3-70B-Instruct & \texttt{meta-llama/Llama-3.3-70B-Instruct} & 0.6 & 0.9 & - \\
\includegraphics[height=1.5ex]{img/model_logos/gemma.png} Gemma-4-31B-IT & \texttt{google/gemma-4-31b-it} & 1.0 & 0.95 & 64 \\
\includegraphics[height=1.5ex]{img/model_logos/openai.png} GPT-5-mini & \texttt{gpt-5-mini-2025-08-07} & 1.0 & 1.0 & - \\
\bottomrule
\end{tabular}
\caption{Model identifiers and their default decoding parameters that we used for our experiments (temperature, top-p, top-k).}
\label{tab:model_params}
\end{table*}

\section{Additional Results}
\label{app:results}

\subsection{Soft F1}

In \Cref{tab:softf1_by_method}, we present the \emph{soft} F1-score results of our models, complementing the \emph{hard} F1-score results in \Cref{tab:hardf1_by_method}. The \emph{soft} F1 score does not penalize differences in category labels, only in span positions.

\begin{table}[t]
  \centering
  % Auto-generated from analysis.ipynb
\small
\setlength{\tabcolsep}{5pt}
\renewcommand{\arraystretch}{1.08}
\begin{tabular}{lrrrr}
\toprule
\textbf{Method} & \textbf{NER} & \textbf{GEC} & \textbf{ESA-MT} & \textbf{CPL} \\
\midrule
\textsc{Tag} & 0.36 & 0.45 & 0.55 & 0.70 \\
\addlinespace[3pt]
\hdashline[0.4pt/2pt]
\addlinespace[3pt]
\textsc{Index} & 0.28 & 0.49 & 0.78 & 1.65 \\
\textsc{Index-Enriched} & 0.77 & 0.64 & 0.59 & 1.51 \\
\addlinespace[3pt]
\hdashline[0.4pt/2pt]
\addlinespace[3pt]
\textsc{Match} & 0.37 & 0.42 & 0.49 & 0.35 \\
\textsc{Match-S} & 0.73 & 0.41 & 0.45 & 2.00 \\
\textsc{Match-Occ} & 0.36 & 0.76 & 0.48 & 0.48 \\
\textsc{Match-Occ-S} & 0.32 & 0.71 & 0.51 & 0.66 \\
\addlinespace[3pt]
\hdashline[0.4pt/2pt]
\addlinespace[3pt]
\textsc{\ourmethod{}} & 0.35 & 1.61 & 0.51 & 0.43 \\
\textsc{\ourmethod{}-S} & 0.65 & 1.52 & 0.40 & 1.80 \\
\textsc{\ourmethod{}-Occ} & 0.36 & 1.71 & 0.51 & 0.62 \\
\textsc{\ourmethod{}-Occ-S} & 0.37 & 1.64 & 0.35 & 0.57 \\
\midrule
\textbf{Overall} & 0.45 & 0.94 & 0.51 & 0.98 \\
\bottomrule
\end{tabular}

  \caption{Mean cross-seed standard deviation of \textbf{hard} F1 in percentage points for each method and task, averaged across open models. The \textbf{Overall} row averages over all method-model configurations within a task.}
  \label{tab:hardf1_seed_std}
\end{table}

\subsection{Token Efficiency Analysis}
\label{app:token_efficiency}

\begin{table}[t]
  \centering
  \small
  \begin{tabular}{lrrr}
\toprule
\textbf{Dataset} & \textbf{input chars} ($n$) & \textbf{spans} ($m$) & \textbf{span len} ($\bar{l}$) \\
\midrule
NER & 112.0 & 1.9 & 8.8 \\
GEC & 867.1 & 14.1 & 4.2 \\
ESA-MT & 250.3 & 2.2 & 13.7 \\
CPL & 536.9 & 12.7 & 6.0 \\
\bottomrule
\end{tabular}

  \caption{Average input text length ($n$, characters), span count per example ($m$), and average span length ($\bar{l}$, characters) from reference annotations.}
  \label{tab:dataset_characteristics}
\end{table}

\Cref{fig:response_overhead} plots the theoretical response length of each method as a function of $n$, $m$, and $\bar{l}$. The curves are derived analytically from our specific output formats:
\begin{itemize}
  \item \textbf{\cattag{}:} $n + m \cdot C_\text{tag}$ --- the response reproduces the full input text with XML-like tags around each span.
  \item \textbf{\catindex{}:} $m \cdot C_\text{idx}$ --- only character-offset--label pairs are output.
  \item \textbf{\catmatch{}:} $C_\text{base} + m \cdot (C_\text{span} + \bar{l})$ --- the JSON response contains a fixed structural wrapper plus per-span fields carrying the span text.
\end{itemize}
The constants are determined by our specific output templates (see \Cref{tab:prompts_overview}), assuming a three-letter category name: $C_\text{tag} = 28$, $C_\text{idx} = 17$, $C_\text{base} = 13$, and $C_\text{span} = 29$ (all in characters).

For comparison, \Cref{tab:dataset_characteristics} reports the average input text length ($n$), span count ($m$), and average span length ($\bar{l}$) measured on reference annotations, aggregated over all sub-datasets within each dataset group.

\subsection{Experiments with Qwen3 Reasoning}
\label{app:reasoning_experiments}
\Cref{tab:softf1_thinking_comparison} summarizes our experiments with Qwen3-8B with reasoning disabled (our default model) vs. Qwen3-8B with reasoning enabled. We note that while in some cases the reasoning helped significantly (e.g. 47\% improvement in soft-F1 score on NER with the \textsc{Index} method), the results occasionally got worse. Our manual inspection of results has shown that the reasoning trace sometimes got very long, exceeding the \texttt{max\_token} limit, even after we increased it to 16,384 (e.g., when the model started to sequentially enumerate all the characters in the input sequence to get their indices), so it was not able to generate the final output.

\subsection{GPT-5-mini Results}
\label{app:gpt_results}

\Cref{tab:hardsoftf1_gpt} presents the hard and soft F1 scores for the proprietary GPT-5-mini model. Note that \textsc{\ourmethod{}} and its variants cannot be applied to GPT-5-mini as it does not expose logits.

\begin{table*}[ht]
  \centering
  % Auto-generated from analysis.ipynb
\small
\setlength{\tabcolsep}{5pt}
\renewcommand{\arraystretch}{1.08}
\begin{tabular*}{\textwidth}{l@{\extracolsep{12pt}}r@{\extracolsep{5pt}}r@{\extracolsep{5pt}}r@{\extracolsep{5pt}}r@{\extracolsep{12pt}}r@{\extracolsep{5pt}}r@{\extracolsep{5pt}}r@{\extracolsep{5pt}}r@{\extracolsep{12pt}}r@{\extracolsep{5pt}}r@{\extracolsep{5pt}}r@{\extracolsep{5pt}}r@{\extracolsep{12pt}}r@{\extracolsep{5pt}}r@{\extracolsep{5pt}}r@{\extracolsep{5pt}}r}
\toprule
 & \multicolumn{4}{c@{\extracolsep{\fill}}}{\textbf{NER}} & \multicolumn{4}{c@{\extracolsep{\fill}}}{\textbf{GEC}} & \multicolumn{4}{c@{\extracolsep{\fill}}}{\textbf{ESA-MT}} & \multicolumn{4}{c}{\textbf{CPL}} \\
\cmidrule(lr){2-5} \cmidrule(lr){6-9} \cmidrule(lr){10-13} \cmidrule(lr){14-17}
\textbf{Method} & \small \shortstack[c]{\includegraphics[height=1.5ex]{img/model_logos/qwen.png}\\ \texttt{8B}} & \small \shortstack[c]{\includegraphics[height=1.5ex]{img/model_logos/mistral.png}\\ \texttt{24B}} & \small \shortstack[c]{\includegraphics[height=1.5ex]{img/model_logos/llama.png}\\ \texttt{70B}} & \small \shortstack[c]{\includegraphics[height=1.5ex]{img/model_logos/gemma.png}\\ \texttt{31B}} & \small \shortstack[c]{\includegraphics[height=1.5ex]{img/model_logos/qwen.png}\\ \texttt{8B}} & \small \shortstack[c]{\includegraphics[height=1.5ex]{img/model_logos/mistral.png}\\ \texttt{24B}} & \small \shortstack[c]{\includegraphics[height=1.5ex]{img/model_logos/llama.png}\\ \texttt{70B}} & \small \shortstack[c]{\includegraphics[height=1.5ex]{img/model_logos/gemma.png}\\ \texttt{31B}} & \small \shortstack[c]{\includegraphics[height=1.5ex]{img/model_logos/qwen.png}\\ \texttt{8B}} & \small \shortstack[c]{\includegraphics[height=1.5ex]{img/model_logos/mistral.png}\\ \texttt{24B}} & \small \shortstack[c]{\includegraphics[height=1.5ex]{img/model_logos/llama.png}\\ \texttt{70B}} & \small \shortstack[c]{\includegraphics[height=1.5ex]{img/model_logos/gemma.png}\\ \texttt{31B}} & \small \shortstack[c]{\includegraphics[height=1.5ex]{img/model_logos/qwen.png}\\ \texttt{8B}} & \small \shortstack[c]{\includegraphics[height=1.5ex]{img/model_logos/mistral.png}\\ \texttt{24B}} & \small \shortstack[c]{\includegraphics[height=1.5ex]{img/model_logos/llama.png}\\ \texttt{70B}} & \small \shortstack[c]{\includegraphics[height=1.5ex]{img/model_logos/gemma.png}\\ \texttt{31B}} \\
\midrule
\textsc{Tag} & \textbf{79.3} & 79.4 & \textbf{87.1} & 92.6 & \textbf{29.4} & \textbf{35.1} & \textbf{37.5} & \textbf{55.9} & 21.3 & 22.2 & 24.7 & \textbf{32.5} & 40.2 & 53.1 & 79.5 & 93.3 \\
\addlinespace[3pt]
\hdashline[0.4pt/2pt]
\addlinespace[3pt]
\textsc{Index} & 26.5 & 32.7 & 32.8 & 57.3 & 11.8 & 10.8 & 10.4 & 10.7 & 18.3 & 17.4 & 16.5 & 18.8 & 34.1 & 40.0 & 36.6 & 57.9 \\
\textsc{Index-Enriched} & 38.9 & 74.2 & 75.2 & 91.4 & 15.6 & 20.1 & 20.5 & 51.5 & 18.1 & 18.8 & 20.4 & 25.8 & 45.4 & \textbf{70.1} & 84.4 & \textbf{98.6} \\
\addlinespace[3pt]
\hdashline[0.4pt/2pt]
\addlinespace[3pt]
\textsc{Match} & 79.2 & \textbf{83.6} & 85.4 & 90.7 & 15.6 & 21.4 & 22.7 & 42.2 & 24.3 & 24.5 & 25.7 & 29.7 & 42.8 & 43.8 & 48.5 & 51.1 \\
\textsc{Match-S} & 77.8 & 80.5 & 85.7 & 88.3 & 14.4 & 19.1 & 22.0 & 41.7 & 24.0 & 25.0 & 25.5 & 27.8 & 42.6 & 41.1 & 47.8 & 39.4 \\
\textsc{Match-Occ} & 78.4 & 83.0 & 85.4 & 91.6 & 19.2 & 24.0 & 23.6 & 45.4 & 23.8 & 25.4 & 25.1 & 30.0 & \textbf{70.1} & 69.1 & \textbf{86.5} & 94.7 \\
\textsc{Match-Occ-S} & 75.8 & 80.3 & 85.5 & 91.1 & 18.2 & 23.3 & 23.8 & 45.3 & 24.5 & \textbf{25.6} & 25.7 & 29.4 & \textbf{70.1} & 67.7 & 83.6 & 93.8 \\
\addlinespace[3pt]
\hdashline[0.4pt/2pt]
\addlinespace[3pt]
\textsc{\ourmethod{}} & 78.7 & 80.6 & 83.4 & 92.5 & 18.7 & 24.7 & 21.2 & 44.4 & 23.9 & 24.8 & 25.7 & 30.7 & 40.1 & 42.9 & 47.4 & 51.1 \\
\textsc{\ourmethod{}-S} & 77.5 & 77.7 & 83.8 & 89.3 & 17.9 & 22.1 & 23.8 & 44.3 & 23.9 & 24.8 & \textbf{25.8} & 27.7 & 41.2 & 40.9 & 46.7 & 39.7 \\
\textsc{\ourmethod{}-Occ} & 78.1 & 79.7 & 82.9 & \textbf{93.3} & 22.9 & 26.0 & 22.8 & 47.8 & 23.7 & 25.3 & 24.5 & 30.6 & 68.6 & 68.9 & 86.3 & 94.7 \\
\textsc{\ourmethod{}-Occ-S} & 75.6 & 77.1 & 83.2 & 92.4 & 21.9 & 25.5 & 25.1 & 48.0 & \textbf{24.8} & 25.0 & 25.6 & 29.4 & 69.6 & 68.0 & 84.4 & 93.7 \\
\bottomrule
\end{tabular*}

  \caption{Per-method \textbf{soft} F1 score in \%. The best score for each model per dataset is bold.}
  \label{tab:softf1_by_method}
\end{table*}

\begin{table*}[ht]
  \centering
  % Auto-generated from analysis.ipynb (Thinking Dual)
\small
\setlength{\tabcolsep}{6pt}
\renewcommand{\arraystretch}{1.08}
\begin{tabular*}{\textwidth}{l@{\extracolsep{8pt}}r@{\extracolsep{5pt}}r@{\extracolsep{9pt}}r@{\extracolsep{5pt}}r@{\extracolsep{8pt}}r@{\extracolsep{5pt}}r@{\extracolsep{9pt}}r@{\extracolsep{5pt}}r@{\extracolsep{8pt}}r@{\extracolsep{5pt}}r@{\extracolsep{9pt}}r@{\extracolsep{5pt}}r@{\extracolsep{8pt}}r@{\extracolsep{5pt}}r@{\extracolsep{9pt}}r@{\extracolsep{5pt}}r}
\toprule
 & \multicolumn{4}{c}{\textbf{NER}} & \multicolumn{4}{c}{\textbf{GEC}} & \multicolumn{4}{c}{\textbf{ESA-MT}} & \multicolumn{4}{c}{\textbf{CPL}} \\
\cmidrule(lr){2-5} \cmidrule(lr){6-9} \cmidrule(lr){10-13} \cmidrule(lr){14-17}
 & \multicolumn{2}{c}{\shortstack[c]{\includegraphics[height=1.5ex]{img/model_logos/qwen.png}\\ \texttt{8B}}} & \multicolumn{2}{c}{\shortstack[c]{\includegraphics[height=1.5ex]{img/model_logos/qwen.png}\\ \texttt{8B-T}}} & \multicolumn{2}{c}{\shortstack[c]{\includegraphics[height=1.5ex]{img/model_logos/qwen.png}\\ \texttt{8B}}} & \multicolumn{2}{c}{\shortstack[c]{\includegraphics[height=1.5ex]{img/model_logos/qwen.png}\\ \texttt{8B-T}}} & \multicolumn{2}{c}{\shortstack[c]{\includegraphics[height=1.5ex]{img/model_logos/qwen.png}\\ \texttt{8B}}} & \multicolumn{2}{c}{\shortstack[c]{\includegraphics[height=1.5ex]{img/model_logos/qwen.png}\\ \texttt{8B-T}}} & \multicolumn{2}{c}{\shortstack[c]{\includegraphics[height=1.5ex]{img/model_logos/qwen.png}\\ \texttt{8B}}} & \multicolumn{2}{c}{\shortstack[c]{\includegraphics[height=1.5ex]{img/model_logos/qwen.png}\\ \texttt{8B-T}}} \\
\cmidrule(lr){2-3} \cmidrule(lr){4-5} \cmidrule(lr){6-7} \cmidrule(lr){8-9} \cmidrule(lr){10-11} \cmidrule(lr){12-13} \cmidrule(lr){14-15} \cmidrule(lr){16-17}
\textbf{Method} & \scriptsize hard & \scriptsize soft & \scriptsize hard & \scriptsize soft & \scriptsize hard & \scriptsize soft & \scriptsize hard & \scriptsize soft & \scriptsize hard & \scriptsize soft & \scriptsize hard & \scriptsize soft & \scriptsize hard & \scriptsize soft & \scriptsize hard & \scriptsize soft \\
\midrule
\textsc{Tag} & 73.3 & \textbf{79.3} & 84.3 & 88.4 & \textbf{27.2} & \textbf{29.4} & 29.3 & \textbf{41.7} & 10.2 & 21.3 & 9.8 & 22.3 & 40.2 & 40.2 & 89.4 & 89.4 \\
\addlinespace[3pt]
\hdashline[0.4pt/2pt]
\addlinespace[3pt]
\textsc{Index} & 19.1 & 26.5 & 69.8 & 74.1 & 10.4 & 11.8 & 6.7 & 8.0 & 7.5 & 18.3 & 7.6 & 17.7 & 34.1 & 34.1 & 67.6 & 67.6 \\
\textsc{Index-Enriched} & 33.7 & 38.9 & 78.5 & 84.3 & 11.0 & 15.6 & 22.9 & 25.1 & 7.5 & 18.1 & 9.9 & 22.7 & 45.4 & 45.4 & \textbf{93.5} & \textbf{93.5} \\
\addlinespace[3pt]
\hdashline[0.4pt/2pt]
\addlinespace[3pt]
\textsc{Match} & \textbf{74.1} & 79.2 & 84.4 & 88.3 & 12.1 & 15.6 & 30.8 & 37.3 & 11.4 & 24.3 & \textbf{10.8} & 25.0 & 42.8 & 42.8 & 51.7 & 51.7 \\
\textsc{Match-S} & 72.8 & 77.8 & \textbf{84.5} & 88.4 & 12.0 & 14.4 & 34.6 & 37.4 & 10.7 & 24.0 & 10.4 & \textbf{25.2} & 42.6 & 42.6 & 51.8 & 51.8 \\
\textsc{Match-Occ} & 73.0 & 78.4 & 76.7 & 81.0 & 15.4 & 19.2 & 32.8 & 38.3 & \textbf{11.9} & 23.8 & 10.2 & 24.2 & \textbf{70.1} & \textbf{70.1} & 75.4 & 75.4 \\
\textsc{Match-Occ-S} & 70.8 & 75.8 & 77.1 & 81.4 & 14.9 & 18.2 & \textbf{35.7} & 38.6 & 11.1 & 24.5 & 10.7 & 24.1 & \textbf{70.1} & \textbf{70.1} & 75.8 & 75.8 \\
\addlinespace[3pt]
\hdashline[0.4pt/2pt]
\addlinespace[3pt]
\textsc{\ourmethod{}} & 73.3 & 78.7 & 84.1 & 88.4 & 17.4 & 18.7 & 33.6 & 36.5 & 11.2 & 23.9 & 9.8 & 22.7 & 40.1 & 40.1 & 51.0 & 51.0 \\
\textsc{\ourmethod{}-S} & 71.9 & 77.5 & 84.4 & \textbf{88.8} & 16.8 & 17.9 & 34.8 & 37.8 & 10.8 & 23.9 & 9.9 & 22.8 & 41.2 & 41.2 & 51.1 & 51.1 \\
\textsc{\ourmethod{}-Occ} & 72.4 & 78.1 & 77.6 & 82.3 & 21.3 & 22.9 & 34.2 & 36.7 & 11.8 & 23.7 & 10.0 & 22.0 & 68.6 & 68.6 & 74.6 & 74.6 \\
\textsc{\ourmethod{}-Occ-S} & 70.1 & 75.6 & 77.7 & 82.3 & 20.4 & 21.9 & 35.2 & 37.9 & 11.2 & \textbf{24.8} & 9.8 & 22.0 & 69.6 & 69.6 & 74.8 & 74.8 \\
\bottomrule
\end{tabular*}

  \caption{Qwen3-8B F1-score (hard, soft) in \% with vs. without reasoning. Qwen3-8B (\texttt{8B}) has reasoning capabilities turned off, while Qwen3-8B-Think (\texttt{8B-T}) has them turned on. The best score for each model per dataset is bold.}
  \label{tab:softf1_thinking_comparison}
\end{table*}

\begin{table*}[ht]
  \centering
  % Auto-generated from analysis.ipynb (Thinking Dual)
\small
\setlength{\tabcolsep}{6pt}
\renewcommand{\arraystretch}{1.08}
\begin{tabular}{lr@{\hspace{5pt}}r@{\hspace{10pt}}r@{\hspace{5pt}}r@{\hspace{10pt}}r@{\hspace{5pt}}r@{\hspace{10pt}}r@{\hspace{5pt}}r}
\toprule
 & \multicolumn{2}{c}{\textbf{NER}} & \multicolumn{2}{c}{\textbf{GEC}} & \multicolumn{2}{c}{\textbf{ESA-MT}} & \multicolumn{2}{c}{\textbf{CPL}} \\
\cmidrule(lr){2-3} \cmidrule(lr){4-5} \cmidrule(lr){6-7} \cmidrule(lr){8-9}
 & \multicolumn{2}{c}{\shortstack[c]{\includegraphics[height=1.5ex]{img/model_logos/openai.png}\\ \texttt{5-m}}} & \multicolumn{2}{c}{\shortstack[c]{\includegraphics[height=1.5ex]{img/model_logos/openai.png}\\ \texttt{5-m}}} & \multicolumn{2}{c}{\shortstack[c]{\includegraphics[height=1.5ex]{img/model_logos/openai.png}\\ \texttt{5-m}}} & \multicolumn{2}{c}{\shortstack[c]{\includegraphics[height=1.5ex]{img/model_logos/openai.png}\\ \texttt{5-m}}} \\
\cmidrule(lr){2-3} \cmidrule(lr){4-5} \cmidrule(lr){6-7} \cmidrule(lr){8-9}
\textbf{Method} & \scriptsize hard & \scriptsize soft & \scriptsize hard & \scriptsize soft & \scriptsize hard & \scriptsize soft & \scriptsize hard & \scriptsize soft \\
\midrule
\textsc{Tag} & 74.3 & 80.3 & \textbf{37.4} & \textbf{41.3} & 12.2 & 27.5 & 73.0 & 73.0 \\
\addlinespace[3pt]
\hdashline[0.4pt/2pt]
\addlinespace[3pt]
\textsc{Index} & 38.3 & 43.3 & 7.7 & 9.2 & 8.0 & 19.0 & 48.8 & 48.8 \\
\textsc{Index-Enriched} & 78.9 & \textbf{85.2} & 27.1 & 32.6 & 11.3 & 24.9 & 83.2 & 83.2 \\
\addlinespace[3pt]
\hdashline[0.4pt/2pt]
\addlinespace[3pt]
\textsc{Match} & 79.0 & 82.3 & 22.4 & 26.3 & 12.1 & 26.7 & 48.4 & 48.4 \\
\textsc{Match-S} & 77.2 & 80.9 & 19.8 & 23.4 & 12.3 & 27.2 & 43.5 & 43.5 \\
\textsc{Match-Occ} & \textbf{79.5} & 83.0 & 26.7 & 30.9 & 12.3 & \textbf{27.8} & 76.4 & 76.4 \\
\textsc{Match-Occ-S} & 76.1 & 79.8 & 25.6 & 29.5 & \textbf{12.5} & 27.6 & \textbf{85.6} & \textbf{85.6} \\
\bottomrule
\end{tabular}

  \caption{GPT-5-mini F1-score (hard, soft) in \% per method and dataset. The best score per dataset is bold.}
  \label{tab:hardsoftf1_gpt}
\end{table*}

\section{Implementation Details}
\label{app:implementation_details}

\subsection{Prompts}
\label{app:prompts}

The overview of the prompts we use for individual methods and tasks is provided in \Cref{tab:prompts_overview}.

\subsection{Methods}
\label{app:methods}
Below, we provide extra implementation details for individual methods.

\paragraph{Tagging} For the \cattag{} strategy, we parse XML-style tags using regular expressions. To map these predictions back to the original text, we estimate the span's location by stripping tags from the preceding output to calculate a running offset. Since models occasionally alter whitespace or hallucinate, we treat this calculated position as a heuristic rather than an absolute truth. If the text at the expected location does not match the tag content, we search the full text for the span and resolve ambiguities by selecting the occurrence closest to our initial estimate. This allows us to recover span boundaries even when the model's output deviates slightly from the input instead of having to discard the output completely.

\paragraph{\textsc{LogitMatch}}
\label{app:algmatch}

The pseudocode for the \textsc{\ourmethod{}} algorithm (\Cref{sec:our_method}) is described in \Cref{alg:logitmatch}.  Note that for presentation purposes, the pseudocode does not account for tokenization issues described  in \Cref{sec:tokenization}. However, we account for these issues in our actual implementation in the project repository. We implement our method as a vLLM \texttt{LogitsProcessor},\footnote{\url{https://docs.vllm.ai/en/latest/design/logits_processors/}} which allows us to combine the method with structured outputs.

% One such issue is quote escaping within JSON, which we do not handle. While technically possible, it would introduce an additional layer of complexity and it is not required for our tasks.

\begin{algorithm*}[hbt]
  % \small
\caption{\ourmethod{} decoding algorithm}
\label{alg:logitmatch}
\begin{algorithmic}[1]
\Require Input $X = [x_1, x_2, \ldots, x_n]$, model $\mathcal{M}$
\State Initialize output sequence $Y$
\While{decoding is not finished}
    \State $t \gets$ next token predicted by $\mathcal{M}(Y)$ \Comment{\textbf{Decoding loop}}
    \State Append $t$ to $Y$
    \If{\texttt{'"text": "'} was decoded} \Comment{\textbf{Select mode}}
        
        \State $V_{\text{select}} \gets \{x \mid x \in X\}$ \Comment{Limit vocabulary to tokens present in input}
        \State $x_i \gets$ sample from $\mathcal{M}(Y)$ constrained to $V_{\text{select}}$
        \State Append $x_i$ to $Y$
        \State $k \gets 1$ \Comment{Initialize relative offset}
        \Loop \Comment{\textbf{Copy mode}}
            \State $V_{\text{copy}} \gets \{x_{i+k}\} \cup \{\texttt{"}\}$ \Comment{Allow next input token or closing quote}
            \State $t_{\text{next}} \gets$ sample from $\mathcal{M}(Y)$ constrained to $V_{\text{copy}}$
            \State Append $t_{\text{next}}$ to $Y$
            
            \If{$t_{\text{next}}$ is \texttt{"}}
                \State \textbf{break} \Comment{Exit copy mode}
            \Else
                \State $k \gets k + 1$ \Comment{Advance relative offset}
            \EndIf
        \EndLoop
    \EndIf
\EndWhile
\end{algorithmic}
\end{algorithm*}

\begin{table*}[t]
\centering
\small
\begin{tabularx}{\textwidth}{l X}
\toprule
\multicolumn{2}{c}{\textbf{Prompt template}} \\
\midrule
\textbf{Structure} & \texttt{[Task description]}\newline\texttt{Output Format: [Format]}\newline\texttt{Labels: [Label list]}\newline\texttt{Examples:}\newline\texttt{1. [Example input] -> [Example output]}\newline\texttt{...}\newline\texttt{[Notes]}\newline\texttt{[Input text]} \\
\midrule
\multicolumn{2}{c}{\textbf{Task definitions}} \\
\midrule
\textbf{NER} & \texttt{Extract named entities (PERSON, ORG, LOC) from the text.} \\
\addlinespace[3pt]
\textbf{GEC} & \texttt{Identify grammatical errors in learner-written text.} \\
\addlinespace[3pt]
\textbf{ESA-MT} & \texttt{Identify translation errors by comparing the translation to the source text.} \\
\addlinespace[3pt]
\textbf{CPL} & \texttt{Find all text spans that match the given pattern queries.} \\
\midrule
\multicolumn{2}{c}{\textbf{Tagging strategies}} \\
\midrule
\textbf{Format} & \texttt{\textless{}entity type="LABEL"\textgreater{}\textless{}/entity\textgreater{}} \\
\addlinespace[3pt]
\textbf{Note} & \texttt{IMPORTANT: Your output needs to include copy of the entire input text, including non-tagged parts. If you are not outputting any tags, you need to copy the input text literally. Surround the specific spans with the XML tags as required.  Do not output any additional explanations or comments, start generating output straight away.} \\
\addlinespace[3pt]
\textbf{Example output} & \texttt{\textless{}entity type="PER"\textgreater{}Lina Berg\textless{}/entity\textgreater{} joined \textless{}entity type="ORG"\textgreater{}AstraTech\textless{}/entity\textgreater{} in \textless{}entity type="LOC"\textgreater{}Stockholm\textless{}/entity\textgreater{} after finishing her studies at \textless{}entity type="ORG"\textgreater{}Northvale University\textless{}/entity\textgreater{}.} \\
\midrule
\multicolumn{2}{c}{\textbf{Matching strategies}} \\
\midrule
\textbf{Format} & \texttt{{[}\{"text": "exact text span from input", "label": "category (PERSON, ORG, LOC)"\}{]}} \\
\addlinespace[3pt]
\textbf{Note} & \texttt{Return a valid JSON array only.} \\
\addlinespace[3pt]
\textbf{Example output} & \texttt{{[} \{"text": "Lina Berg", "label": "PER"\}, \{"text": "AstraTech", "label": "ORG"\}, \{"text": "Stockholm", "label": "LOC"\}, \{"text": "Northvale University", "label": "ORG"\} {]}} \\[2pt]
\hdashline[0.4pt/2pt]
\addlinespace[2pt]
\multicolumn{2}{l}{\textit{For Match-Occ}} \\
\addlinespace[3pt]
\textbf{Format} & \texttt{{[}\{"text": "exact span from input", "label": "category (PERSON, ORG, LOC)", "occurrence": "which occurrence (1, 2, 3...)"\}{]}} \\
\addlinespace[3pt]
\textbf{Example output} & \texttt{{[} \{"text": "Lina Berg", "label": "PER", "occurrence": 1\}, \{"text": "AstraTech", "label": "ORG", "occurrence": 1\}, \{"text": "Stockholm", "label": "LOC", "occurrence": 1\}, \{"text": "Northvale University", "label": "ORG", "occurrence": 1\} {]}} \\
\midrule
\multicolumn{2}{c}{\textbf{Indexing strategies}} \\
\midrule
\textbf{Format} & \texttt{{[}start:end{]} = LABEL} \\
\addlinespace[3pt]
\textbf{Note} & \texttt{IMPORTANT: Character positions are 0-indexed. - First character is at position 0 - Spaces count as characters - start is inclusive, end is exclusive} \\
\addlinespace[3pt]
\textbf{Example output} & \texttt{{[}0:9{]} = PER {[}17:26{]} = ORG {[}30:39{]} = LOC {[}71:91{]} = ORG} \\
\bottomrule
\end{tabularx}
\caption{Overview of prompts used for each strategy.}
\label{tab:prompts_overview}
\end{table*}

\subsection{Data Preprocessing}
\label{app:datasets}
We made several adjustments to the MultiGEC dataset \cite{masciolini2025better} that we used for the GEC task:

\begin{itemize}
  \item The original MultiGEC dataset contains texts in 12 languages in total. We initially wanted to include all the languages in our experiments. However, we found that for the majority of languages, the inputs (text with errors) and the outputs (corrected texts) are not aligned. Only the \emph{English} subset (Write \& Improve corpus) contained the list of edits between the original and corrected sentences, which corresponds directly to the span labeling task. Therefore, we limited our dataset only to the English subset.
  \item Since we focus mainly on span identification, we ignore the corrected text for the replacement and missing edits. For the same reason, we also use only the high-level categories (replacement, missing, and unnecessary) and ignore the more fine-grained categories.
  \item The original MultiGEC dataset represents missing spans (label
\texttt{M}) as zero-length spans, where the start and end index
are equal and indicate the insertion point. Since matching-based
methods cannot refer to an empty span, we instruct the model in
the \textsc{Match} and \textsc{Match-Occ} prompts to output the
token immediately preceding the insertion point instead. In
post-processing, we advance the predicted start index past the
output token to recover the insertion point. The evaluation
metric (\Cref{sec:f1}) treats each
zero-length span as contributing a unit weight to the relevant
F1 denominator, with a zero-length predicted span counting as a
match for a zero-length gold span only when their positions
coincide and (under hard matching) their labels agree. The
tagging and indexing strategies do not receive this instruction and their outputs are parsed as-is.
\end{itemize}

\end{document}